\documentclass{article}
\pdfoutput=1
\usepackage{microtype}
\usepackage{graphicx}
\usepackage{subfigure}
\usepackage{booktabs} 

\usepackage{hyperref}

\usepackage[accepted]{icml2023}

\usepackage{amsmath}
\usepackage{amssymb}
\usepackage{mathtools}
\usepackage{amsthm}

\usepackage[capitalize,noabbrev]{cleveref}

\theoremstyle{plain}
\newtheorem{theorem}{Theorem}[section]

\theoremstyle{definition}

\theoremstyle{remark}

\makeatletter
\newtheorem*{rep@theorem}{\rep@title}
\newcommand{\newreptheorem}[2]{%
\newenvironment{rep#1}[1]{%
 \def\rep@title{#2 \ref{##1}}%
 \begin{rep@theorem}}%
 {\end{rep@theorem}}}
\makeatother
\newreptheorem{theorem}{Theorem}
\newreptheorem{lemma}{Lemma}

\usepackage{amsmath,amsfonts,bm}

\def\eqref#1{equation~\ref{#1}}

\def\1{\bm{1}}

\def\rvp{{\mathbf{p}}}

\def\rvx{{\mathbf{x}}}
\def\rvy{{\mathbf{y}}}

\def\vmu{{\bm{\mu}}}
\def\vtheta{{\bm{\theta}}}

\def\vb{{\bm{b}}}

\def\vp{{\bm{p}}}

\def\vx{{\bm{x}}}
\def\vy{{\bm{y}}}

\def\mI{{\bm{I}}}

\DeclareMathAlphabet{\mathsfit}{\encodingdefault}{\sfdefault}{m}{sl}
\SetMathAlphabet{\mathsfit}{bold}{\encodingdefault}{\sfdefault}{bx}{n}

\newcommand{\E}{\mathbb{E}}

\newcommand{\KL}{D_{\mathrm{KL}}}

\newcommand{\Cov}{\mathrm{Cov}}

\usepackage[textsize=tiny]{todonotes}

\icmltitlerunning{Uncertainty Estimation by Fisher Information-based Evidential Deep Learning}

\begin{document}

\twocolumn[
\icmltitle{Uncertainty Estimation by Fisher Information-based Evidential Deep Learning}



\icmlsetsymbol{equal}{*}

\begin{icmlauthorlist}
\icmlauthor{Danruo Deng}{cuhk,mixr}
\icmlauthor{Guangyong Chen}{zjl}
\icmlauthor{Yang Yu}{cuhk,mixr}
\icmlauthor{Furui Liu}{zjl}
\icmlauthor{Pheng-Ann Heng}{cuhk,mixr}
\end{icmlauthorlist}

\icmlaffiliation{cuhk}{Department of Computer Science and Engineering, The Chinese University of Hong Kong}
\icmlaffiliation{zjl}{Zhejiang Lab}
\icmlaffiliation{mixr}{Institute of Medical Intelligence and XR, The Chinese University of Hong Kong}

\icmlcorrespondingauthor{Guangyong Chen}{gychen@zhejianglab.com}

\icmlkeywords{Uncertainty Estimation, Machine Learning, ICML}

\vskip 0.3in
]



\printAffiliationsAndNotice{}  

\begin{abstract}
Uncertainty estimation is a key factor that makes deep learning reliable in practical applications. Recently proposed evidential neural networks explicitly account for different uncertainties by treating the network's outputs as evidence to parameterize the Dirichlet distribution, and achieve impressive performance in uncertainty estimation. However, for high data uncertainty samples but annotated with the one-hot label, the evidence-learning process for those mislabeled classes is over-penalized and remains hindered. To address this problem, we propose a novel method, \textit{Fisher Information-based Evidential Deep Learning} ($\mathcal{I}$-EDL). In particular, we introduce Fisher Information Matrix (FIM) to measure the informativeness of evidence carried by each sample, according to which we can dynamically reweight the objective loss terms to make the network more focus on the representation learning of uncertain classes. The generalization ability of our network is further improved by optimizing the PAC-Bayesian bound. As demonstrated empirically, our proposed method consistently outperforms traditional EDL-related algorithms in multiple uncertainty estimation tasks, especially in the more challenging few-shot classification settings. 
\end{abstract}

\section{Introduction}
\label{sec_intro}

Uncertainty estimation is crucial not only for safety decision-making in high-risk domains such as medical image analysis \cite{seebock2019exploiting, nair2020exploring} and autonomous vehicle control \cite{feng2018towards, choi2019gaussian}, but also in the general fields where data is highly heterogeneous or scarcely annotated \cite{gal2017deep, ablain2019uncertainty}. Predictive uncertainty is quite diverse and can be divided into \textit{data uncertainty, model uncertainty}, and \textit{distributional uncertainty} \cite{Gal2016UncertaintyID, malinin2018predictive}. Data uncertainty, or aleatoric uncertainty, is caused by the natural complexity of the data, such as class overlap and label noise. Model uncertainty, or epistemic uncertainty, measures the uncertainty in estimating model parameters given training data. Model and data uncertainty are sometimes referred to as reducible and irreducible uncertainty, respectively, since model uncertainty can be reduced with more training data, while data uncertainty cannot. Distributional uncertainty arises from a distribution mismatch between the training and test distributions, i.e., the test data is out-of-distribution (OOD) \cite{quinonero2008dataset}. Quantifying the different uncertainty is obviously a key factor that makes deep learning reliable.

\textit{Softmax} is the most widely used normalization function that maps the continuous activations of the output layer to a probability distribution. However, \textit{Softmax} is notorious for inflating the probabilities of predicted classes \cite{szegedy2016rethinking, guo2017calibration, wilson2020bayesian}. Some methods calibrate network predictions to improve the reliability of uncertainty estimation \cite{guo2017calibration, liang2018enhancing}. However, these methods can still not distinguish between different types of uncertainty, which seriously limits the practical usage of deep learning in challenging domains without enough training samples. Recently notable progress has been made in estimating uncertainty in DNNs. One class of methods stems from Bayesian neural networks, which quantify uncertainty by learning a posterior over weights \cite{gal2016dropout, ritter2018scalable, kristiadi2022being}. Other methods often combine predictions from several independently trained networks to estimate statistics for class probability distributions \cite{lakshminarayanan2017simple, zaidi2021neural}. However, these approaches still cannot distinguish distributional uncertainty from other uncertainties \cite{malinin2018predictive}.

\begin{figure}[t]
    \centering
    \includegraphics[width=0.4\textwidth]{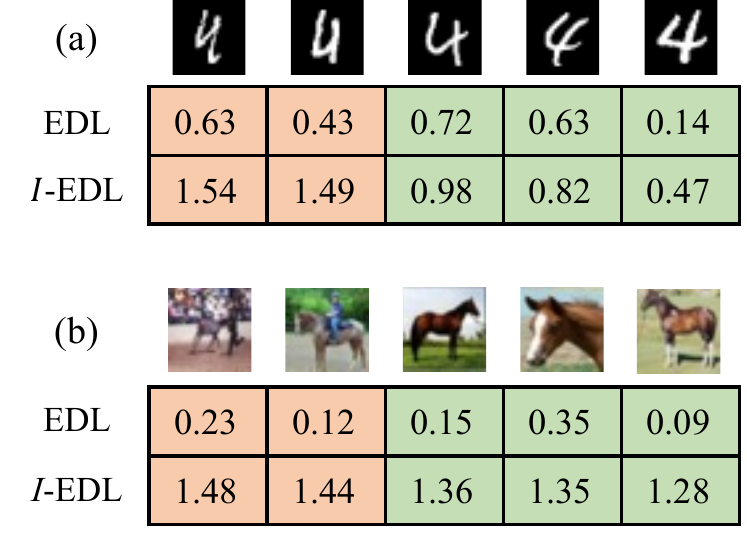}
    \caption{Data uncertainty for (a) digit ``4" in MNIST, (b) ``horse" in CIFAR10. $\mathcal{I}$-EDL has the ability to distinguish between hard samples (orange) and easy samples (green), but EDL cannot.}
    \label{fig_case}
\end{figure}

To address this limitation, Dirichlet-based uncertainty models quantify different types of uncertainty by modeling the output as the concentration parameters of a Dirichlet distribution \cite{malinin2018predictive, malinin2019reverse, nandy2020towards, charpentier2020posterior}. Evidential deep learning (EDL) \cite{sensoy2018evidential} adopts Dirichlet distribution and treats the output as evidence to quantify belief mass and uncertainty by jointly considering the Dempster–Shafer Theory of Evidence (DST) \cite{dempster1968generalization, shafer1976mathematical} and subjective logic (SL) \cite{josang2016subjective}. The evidential network proposed by EDL can be represented as a probabilistic graphical model, where the observed labels $\vy$ are generated from the Dirichlet distribution with its parameter $\boldsymbol{\alpha}$ calculated by passing the input sample $\vx$ through the network. EDL, which learns optimal parameters by maximizing the expected likelihood of the observed labels, shows an impressive performance in uncertainty quantification and is widely used in various applications, such as graph neural networks \cite{zhao2020uncertainty}, open set recognition (OSR) \cite{bao2021evidential, bao2022opental}, molecular property prediction \cite{soleimany2021evidential}, meta-learning \cite{pandey2022multidimensional}, and active learning \cite{hemmer2022deal}. However, for samples with high data uncertainty but annotated with one-hot vectors, the learning process of evidence for those mislabeled classes is over-penalized and remains hindered. Data uncertainty is ubiquitous in real-world applications, such as the indistinguishable ``4" and ``9", ``1" and ``7" in the MNIST dataset, images containing multiple objects in ImageNet, and the unavoidable noise labels in almost all datasets. Figure \ref{fig_case} illustrates that EDL cannot correctly distinguish MNIST and CIFAR10 image samples with different data uncertainties. Although EDL can model different types of uncertainty, its training process will underestimate the irreducible data uncertainty, thereby reducing the availability of uncertainty estimation.

\begin{figure}[t]
    \centering
    \includegraphics[width=0.35\textwidth]{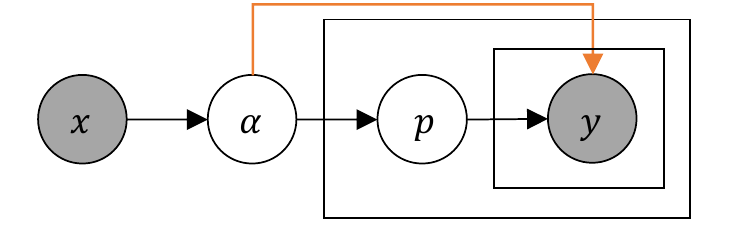}
    \caption{Graphical model representation of $\mathcal{I}$-EDL. The generative process of observed labels $\vy$ is not only related to the predicted categorical probability $\rvp \sim Dir(\boldsymbol{\alpha})$, but also to the parameters $\boldsymbol{\alpha}$ of the Dirichlet distribution.}
    \label{fig_graphical_model}
\end{figure}

In this paper, we propose a simple and novel method, \textit{Fisher Information-based Evidential Deep Learning} ($\mathcal{I}$-EDL), to weigh the importance of different classes for each training sample. In particular, we introduce Fisher Information Matrix (FIM) to measure the information amount of evidence carried by the categorical probabilities $\vp$ for each sample $\vx$. According to the derivation of Dirichlet distribution's FIM, we found that the higher the evidence for a certain class label, the less corresponding information. Thus, we can set up a Gaussian distribution to help generate observed labels $\vy$ by setting a larger variance for these less informative classes. As shown in Figure \ref{fig_graphical_model}, the generative process of observed labels $\vy$ is not only related to the predicted categorical probability $\rvp$, but also to the parameters $\boldsymbol{\alpha}$ of the Dirichlet distribution. Take the example of an informative image containing both a dog and a cat. The evidence for the dog and cat classes should be high if the neural network learns correctly. Our proposed generative model allows the observed label to be either dog or cat while retaining evidence for the other label, when the observed labels are set as one-hot vectors. From the perspective of optimization, compared with classical EDL, our proposed model encourages the network to focus more on the accuracy of classifying uncertain classes during the training process. From a generative model perspective, compared to classical EDL which only considers a single observation $\vy$, the FIM we introduce can be seen as a type of ground truth distribution. These improvements help enhance the classification performance and uncertainty estimation of the evidential neural networks.
Finally, we further improve the generalization ability of our network by optimizing the PAC-Bayesian bound. 

To our knowledge, we are the first to explicitly leverage evidence to improve uncertainty reliability by dynamically weighing objective loss terms. Our main contributions can be summarized as follows: 
(1) We propose a novel method to quantify uncertainty by combining Fisher information and evidential neural networks. (2) We introduce PAC-Bayesian bound to further improve the generalization ability. (3) Our proposed method achieves excellent empirical performance in confidence evaluation and OOD detection, especially in the more challenging few-shot classification setting.

\section{Preliminary}
\label{sec_pre}

Evidential deep learning (EDL) \citep{sensoy2018evidential} models the neural network by using the theory of subjective logic (SL) \citep{josang1997artificial, josang2016subjective}, a type of probabilistic logic that explicitly takes into account epistemic uncertainty and source trust. The opinion representation used to represent beliefs in SL provides greater expressive power than Boolean truths and probabilities, as it can express \textit{``I don't know"} as an opinion on the truth of possible states. The concept of beliefs is derived from the Dempster–Shafer Theory of Evidence (DST), which is a generalization of the Bayesian theory to subjective probabilities \citep{dempster1968generalization, shafer1976mathematical}. The main idea behind DST is to abandon the additivity principle of probability theory, which means that the sum of belief masses can be less than 1, and the remainder is supplemented by uncertainty mass, i.e., lack of evidence about the truth of state values. 

More specifically, considering a state space consisting of $K$ mutually exclusive singletons (e.g., class labels), SL provides a belief mass $b_k$ for each singleton $k \in [K]$ and an overall uncertainty mass $u$. The $K+1$ mass values satisfy
$
u + \sum_{k=1}^{K} b_k= 1,
$
where $u \geq 0, b_k \geq 0, \forall k \in [K]$. Belief mass depends on the evidence for the corresponding singleton, which measures the amount of support collected from data. In the absence of evidence, the belief for each singleton is $0$, and uncertainty is $1$. Conversely, an infinite amount of evidence leaves no room for uncertainty, yielding belief masses that sum to 1. SL formalizes the belief assignment of DST as a Dirichlet distribution with concentration parameters $\alpha_{k} = e_{k} + 1$, where $e_k$ denotes the derived evidence for the $k$-th singleton. That is, the belief $b_k$ and the uncertainty $u$ can easily be derived from the parameters of the corresponding Dirichlet distribution by using 
$$
b_k = \frac{\alpha_{k} - 1}{\alpha_{0}}, \quad u = \frac{K}{\alpha_{0}}, 
$$
where $\alpha_{0} = \sum_{k=1}^{K} \alpha_{k}$ is referred to as the precision of the Dirichlet distribution. Higher values of $\alpha_{0}$ lead to sharper, more confident distributions.

The Dirichlet distribution is the conjugate prior of the categorical distribution. It is parameterized by the concentration parameters $\boldsymbol{\alpha} = [\alpha_{1}, \cdots, \alpha_{K}]$, $\forall \alpha_{k}>0$, defined as:
$$
Dir(\vp|\boldsymbol{\alpha})=\frac{\Gamma\left(\alpha_{0}\right)}{\prod_{k=1}^{K} \Gamma\left(\alpha_{k}\right)} \prod_{k=1}^{K} p_{k}^{\alpha_{k}-1}, \alpha_{0}=\sum_{k=1}^{K} \alpha_{k}
$$
where $\vp \in \Delta^{K-1}$, and $\Gamma(\cdot)$ is the \textit{gamma} function. Given an opinion, the expected probability of the $k$-th singleton is equivalent to the mean of the corresponding Dirichlet distribution, calculated as 
$$
\hat{p}_k = \frac{\alpha_{k}}{\alpha_{0}} = \frac{e_{k} + 1}{\sum_{c=1}^{K} e_c + K}. 
$$
When the evidence for one of the $K$ attributes is observed, the corresponding Dirichlet parameter is incremented to update the Dirichlet distribution with the new observation. 

EDL encourages neural networks to formalize the multiple opinions for the classification of a given sample as a Dirichlet distribution \citep{sensoy2018evidential}. Compared to standard neural networks, EDL replaces the last softmax layer with an activation layer, e.g., ReLU or Softplus, to obtain a non-negative output, which is the evidence vector used to parameterize the Dirichlet distribution. Standard DNNs for classification with a softmax output function can be viewed as predicting the expected classification distribution of EDL with an exponential output function. This means it is not sensitive to arbitrary scaling of $\alpha_k$. Compared with the standard neural network classifiers that directly output the classification probability distribution of each sample, evidential neural networks obtain the density of classification probability assignments by parameterizing the Dirichlet distribution. Therefore, EDL models second-order probability and uncertainty \citep{josang2016subjective}, which can use the properties of Dirichlet distribution to distinguish different types of uncertainties, but classical EDL hinders the learning of evidence, especially for samples with high data uncertainty.
\section{Method}
\label{sec_method}

\subsection{Generative Model of Evidential Network}
\label{method_intro}

Evidential neural networks are typically trained using a combination of the expected mean squared error (MSE) and a Kullback-Leibler (KL) divergence term as a loss function, where the KL term penalizes evidence for classes that do not fit the training data. Note that, MSE performs best compared with the cross-entropy loss and the negative log marginal likelihood as empirically demonstrated in Sensoy et al. \yrcite{sensoy2018evidential}. Let $f_{\vtheta}: \mathbb{R}^{d} \rightarrow \mathbb{R}^{K}_{+}$ denotes the evidential neural network. Given a sample $(\vx, \vy)$, where $\vy$ is the one-hot encoded ground-truth class of observation $\vx$, the loss function of EDL is expressed as:
$$
\begin{aligned}
    \mathcal{L}(\vtheta) =  \ & \E_{\vp \sim Dir(\boldsymbol{\alpha})} \left[ (\vy - \vp)^T (\vy - \vp) \right] \\
    & + \lambda \KL(Dir(\vp | \hat{\boldsymbol{\alpha}}) \Vert Dir(\vp | \boldsymbol{1})),
\end{aligned}
$$
where $\boldsymbol{\alpha} = f_{\vtheta}(\vx) + 1, \hat{\boldsymbol{\alpha}} = \boldsymbol{\alpha} \odot(1 - \vy) + \vy, \lambda \geq 0$,  $\boldsymbol{1} = [1; \cdots; 1] \in \mathbb{R}^{K}$, and $Dir(\vp | \boldsymbol{1})$ denotes the uniform Dirichlet distribution. 

Actually, when trained with MSE loss, the evidential network proposed by EDL can be understood as a new probabilistic graphical model. Specifically, let $\rvx, \rvy$ denote random variables whose unknown probability distribution generates inputs $\vx \in \mathbb{R}^{d}$ and labels $\vy \in \mathbb{R}^{K}$, respectively.
EDL \cite{sensoy2018evidential} supposes the observed labels $\vy$ were drawn \textit{i.i.d.} from an isotropic Gaussian distribution, i.e. 
$$
\vy \sim \mathcal{N} (\vp, \sigma^2\mI), 
$$
where $\vp \sim Dir(f_{\vtheta}(\vx) + 1)$. Then, training evidential neural networks by minimizing the expected MSE can be viewed as learning model parameters that maximize the expected likelihood of the observed labels. Since the observed labels $\vy$ are one-hot encoded, and the Gaussian distribution of the generated labels is isotropic, i.e. each class is treated independently with the same variance, for high data uncertainty samples, the learning process of evidence for those mislabeled classes is over-penalized and remains hindered. This results in the total amount of evidence being underestimated and make the network overfitting. Furthermore, recently proposed work \cite{bengs2022pitfalls} also argues that classical EDL does not incentivize learners to faithfully predict their epistemic uncertainty due to its sensitivity to the regularization parameter $\lambda$.

\subsection{Fisher Information-based Evidential Network}
\label{method_iedl}

In this work, based on the fact that the information of each class carried in categorical probabilities $\vp$ is different, we argue that the generation of each class for a specific sample should not be isotropic. Intuitively, a certain class label with higher evidence is allowed to have a larger variance, so that the evidence for missing labels can be preserved while maximizing the likelihood of the observed labels. The Fisher information matrix (FIM) is chosen to measure the amount of information that the categorical probabilities $\vp$ carry about the concentration parameters $\boldsymbol{\alpha}$ of a Dirichlet distribution that models $\vp$. Formally, the FIM is defined as:
\begin{equation}
\label{eq:fisher}
\mathcal{I}(\boldsymbol{\alpha}) = \E_{Dir(\vp|\boldsymbol{\alpha})} \left[ \frac{\partial \ell}{\partial \boldsymbol{\alpha}} \frac{\partial \ell}{\partial \boldsymbol{\alpha}^{T}}\right],
\end{equation}
where $\ell = \log Dir(\vp|\boldsymbol{\alpha})$ is the log-likelihood function. Under weak conditions (see Lemma 5.3 in Lehmann \& Casella \yrcite{lehmann2006theory}), the FIM can be expressed as $\mathcal{I}(\boldsymbol{\alpha}) = \E_{Dir(\vp|\boldsymbol{\alpha})} \left[ - \partial^2 \ell /  \partial \boldsymbol{\alpha} \boldsymbol{\alpha}^{T} \right]$. After applying a series of derivation steps (see Appendix \ref{app_proof_fisher} for details), $\mathcal{I}(\boldsymbol{\alpha})$ can be simplified to:
$$
\label{fim_result}
\mathcal{I}(\boldsymbol{\alpha}) = \text{diag}([\psi^{(1)}(\alpha_{1}), \cdots, \psi^{(1)}(\alpha_{K})])- \psi^{(1)}(\alpha_{0}) \boldsymbol{1}\boldsymbol{1}^T,
$$
where $\psi^{(1)}(\cdot)$ denotes the \textit{trigamma} function, defined as $\psi^{(1)}(x) = d \psi(x) / dx = d^2 \ln \Gamma(x) / dx^2 $. Since $\psi^{(1)}(x)$ is a monotonically decreasing function when $x > 0$, the class label with higher evidence corresponds to less Fisher information. Hence, we use the inverse of the FIM ($\mathcal{I}(\boldsymbol{\alpha})^{-1}$) as the variance of the generative distribution of $\vy$. 

Thus, we assume that the target variable $\vy$ follows a multivariate Gaussian distribution with the following closed form: 
$$
\vy \sim \mathcal{N}(\vp , \sigma^2 \mathcal{I}(\boldsymbol{\alpha})^{-1}),
$$
where $\boldsymbol{\alpha} = f_{\vtheta}(\vx) + 1, \vp \sim Dir(\boldsymbol{\alpha})$, $\sigma^2$ is the scalar used to adjust covariance value, $\mathcal{I}(\boldsymbol{\alpha})$ is the FIM of $Dir(\boldsymbol{\alpha})$, defined as Eq.(\ref{eq:fisher}). In MLE, we aim to learn model parameters $\vtheta$ that maximize the marginal likelihood obtained by integrating the class probabilities, i.e.
\begin{equation}
\label{eq:obj_mle}
\max_{\vtheta} \ \E_{(\vx, \vy) \sim \mathcal{P}} \left[\log \E_{\vp \sim Dir(\boldsymbol{\alpha})} [ \mathcal{N}(\vy|\vp, \sigma^2\mathcal{I}(\boldsymbol{\alpha})^{-1}) ]\right].
\end{equation}
Due to the concavity of the log function, by Jensen’s inequality, our objective of Eq.(\ref{eq:obj_mle}) can be achieved by minimizing the expected negative log-likelihood loss function:
\begin{equation}
\begin{aligned}
\label{obj_func_exp}
\min_{\vtheta} \quad &\E_{(\vx, \vy) \sim \mathcal{P}} \E_{\vp \sim Dir(\boldsymbol{\alpha})} \left[- \log p(\vy|\vp, \boldsymbol{\alpha}, \sigma^2)\right]. \\
\text{s.t.} \quad &\boldsymbol{\alpha} = f_{\vtheta}(\vx) + 1 \\
&\mathcal{I}(\boldsymbol{\alpha}) = \E_{Dir(\vp|\boldsymbol{\alpha})} \left[- \frac{\partial^2 \log Dir(\vp|\boldsymbol{\alpha})}{\partial \boldsymbol{\alpha} \boldsymbol{\alpha}^{T}}\right] \\
&p(\vy|\vp, \boldsymbol{\alpha}, \sigma^2) = \mathcal{N}(\vy|\vp, \sigma^2 \mathcal{I}(\boldsymbol{\alpha})^{-1}) \\
\end{aligned}
\end{equation}

\subsection{Learning with PAC-Bayesian Bound}

Since the PAC-Bayesian theory \cite{mcallester1999some} provides data-driven generalization bounds computed on the training set and are simultaneously valid for all posteriors on network parameters, it is often used as a criterion for model selection or as an inspiration for learning algorithm conception. In the PAC-Bayes setting, it assumes that the predictor $f_{\vtheta}$ has prior knowledge of the hypothesis space $\boldsymbol{\Theta}$ in the form of a prior distribution $\pi$. After the training dataset $\mathcal{D}$ is fed to the predictor, the prior is updated to a posterior distribution $\rho$. The full bound theorem is restated below, derived from the theorems in Germain et al. \yrcite{germain2009pac}, Alquier et al. \yrcite{alquier2016properties}, Masegosa \yrcite{masegosa2020learning}, and we give the proof in appendix \ref{app_proof_pac} for completeness.

\begin{theorem}[\cite{germain2009pac, alquier2016properties, masegosa2020learning}]
\label{pac}
Given a data distribution $\mathcal{P}$ over $\mathcal{X} \times \mathcal{Y}$, a hypothesis set $\vtheta$, a prior distribution $\pi$ over $\boldsymbol{\Theta}$, for any $\delta \in (0,1]$, and $\lambda > 0$, with probability at least $1-\delta$ over samples $\mathcal{D} \sim \mathcal{P}^{n}$, we have for all posterior $\rho$,
$$
\begin{aligned}
\E_{\rho(\vtheta)} [\mathcal{L}(\vtheta)] \leq & \ \E_{\rho(\vtheta)} [\hat{\mathcal{L}}_{\mathcal{D}}(\vtheta) ] \\ 
& + \frac{1}{\lambda}\left[\KL(\rho \Vert \pi) + \log \frac{1}{\delta} + \Psi_{\mathcal{P}, \pi}(\lambda, n) \right],
\end{aligned}
$$
where $\Psi_{\mathcal{P}, \pi}(\lambda, n) = \log \E_{\pi(\vtheta)} \E_{\mathcal{D} \sim \mathcal{P}^{n}}\left[e^{\lambda(\mathcal{L}(\vtheta)-\hat{\mathcal{L}}_{\mathcal{D}}(\vtheta))}\right] $.
\end{theorem}

In this paper, we treat $Dir(\vp|\boldsymbol{\alpha})$ as the posterior distribution, and the prior as $Dir(\vp|\vmu)$, where $\vmu$ is set to $\beta \gg 1$ for the corresponding class and $1$ for all other class. Given training set $\mathcal{D} = \{(\vx_i, \vy_i)\}_{i=1}^{N}$, $\pi, \lambda$ and $\delta$, by the Theorem \ref{pac}, the upper bound of Eq.(\ref{obj_func_exp}) can be expressed as
\begin{equation}
\label{obj_func_emp}
\frac{1}{N} \sum_{i=1}^{N} \mathcal{L}_{i}(\vtheta) + \frac{1}{\lambda} \KL(Dir(\vp_i | \boldsymbol{\alpha}_i) \Vert Dir(\vp_i|\vmu_i) ),
\end{equation}
where $\mathcal{L}_{i}(\vtheta) = \E_{Dir(\vp_i|\boldsymbol{\alpha}_i)} \left[ - \log \mathcal{N}(\vy_i|\vp_i, \sigma^2\mathcal{I}(\boldsymbol{\alpha}_i)^{-1})\right]$, $\boldsymbol{\alpha}_i = f_{\vtheta}(\vx_i) + 1$ and $\vp_i \sim Dir(\boldsymbol{\alpha}_i)$.

The first term in Eq.(\ref{obj_func_emp}) can be reformulated as a trade-off between the expected FIM-weighted MSE ($\mathcal{I}$-MSE) and a penalty of negative log determinant of the FIM ($|\mathcal{I}|$), since
$$
\mathcal{L}_{i}(\vtheta) 
\propto \underbrace{\E \left[ (\vy_i - \vp_i)^T \mathcal{I}(\boldsymbol{\alpha}_i) (\vy_i - \vp_i) \right]}_{\mathcal{L}_{i}^{\mathcal{I}\text{-MSE}}} - \sigma^2 \underbrace{\log |\mathcal{I}(\boldsymbol{\alpha}_i)|}_{\mathcal{L}_{i}^{|\mathcal{I}|}}.
$$
Among them, since the class label with less evidence corresponds to the larger Fisher information, minimizing $\mathcal{L}_{i}^{\mathcal{I}\text{-MSE}}$ can be viewed as: for a certain class with low evidence, we expect its corresponding prediction probability to be more accurate, regardless of whether this class is the ground-truth label or not. The penalty for adding $-\mathcal{L}_{i}^{|\mathcal{I}|}$ can be seen as avoiding overconfidence caused by excessive evidence. It is worth noting that these loss terms are not only related to classes but also related to samples.
Thus, the FIM-weighted term can be considered as an adaptive weight that can self-adjust the corresponding MSE loss based on the information of each class contained in the sample.

For the second term in Eq.(\ref{obj_func_emp}), $\vmu_i$ varies with sample labels. To simplify this Kullback-Leibler (KL) divergence, we set $\hat{\boldsymbol{\alpha}}_i = \boldsymbol{\alpha}_i \odot(1 - \vy_i) + \vy_i$ to remove the predicted concentration parameter of the true label corresponding to the sample $\vx_i$ following Sensoy et al. \yrcite{sensoy2018evidential}, thereby the KL term is converted to 
$$
\mathcal{L}_{i}^{\text{KL}} = \KL(Dir(\vp_i | \hat{\boldsymbol{\alpha}}_i) \Vert Dir(\vp_i | \boldsymbol{1})).
$$

Finally, the objective function Eq.(\ref{obj_func_emp}) can be reformulated as
\begin{equation}
\label{obj_final}
\min_{\vtheta} \frac{1}{N} \sum_{i=1}^{N} \mathcal{L}_{i}^{\mathcal{I}\text{-MSE}} -\lambda_1 \mathcal{L}_{i}^{|\mathcal{I}|} + \lambda_2 \mathcal{L}_{i}^{\text{KL}},
\end{equation}
where 
$$
\begin{aligned}
&\mathcal{L}_{i}^{\mathcal{I}\text{-MSE}} 
= \sum_{j=1}^{K} \left( (y_{i j} - \frac{\alpha_{i j}}{\alpha_{i 0}} )^2 + \frac{\alpha_{i j}( \alpha_{i 0} - \alpha_{i j})}{\alpha_{i 0}^2(\alpha_{i 0} + 1)} \right) \psi^{(1)}(\alpha_{i j}), \\
\end{aligned}
$$
$$
\begin{aligned}
&\mathcal{L}_{i}^{|\mathcal{I}|}
= \sum_{j=1}^{K} \log  \psi^{(1)}(\alpha_{i j}) + \log \left(1 - \sum_{j=1}^{K} \frac{ \psi^{(1)}(\alpha_{i 0})}{ \psi^{(1)}(\alpha_{i j})} \right),
\end{aligned}
$$
$$
\begin{aligned}
\label{obj_final_kl}
\mathcal{L}_{i}^{\text{KL}} 
= & \log \Gamma(\sum_{j=1}^{K} \hat{\alpha}_{i j}) - \log \Gamma(K) - \sum_{j=1}^{K} \log \Gamma(\hat{\alpha}_{i j}) \\
& + \sum_{j=1}^{K}(\hat{\alpha}_{i j} - 1) \left[ \psi(\hat{\alpha}_{i j}) - \psi( \sum_{k=1}^{K} \hat{\alpha}_{i k}) \right],
\end{aligned}
$$
and $\lambda_1, \lambda_2 \geq 0$. The detailed derivation steps are left in Appendix \ref{app_proof_obj}. The complete pseudo-code of our method is outlined in Algorithm \ref{alg_train}. Actually, classical EDL can be viewed as a degenerate version of $\mathcal{I}$-EDL without $\mathcal{L}_{i}^{|\mathcal{I}|}$ and with $\mathcal{L}_{i}^{\text{MSE}}$ instead of $\mathcal{L}_{i}^{\mathcal{I}\text{-MSE}}$. Furthermore, we introduce the KL term from the PAC-Bayesian bound to make it more reasonable and interpretable. A detailed comparison of $\mathcal{I}$-EDL and classical EDL is given in Table \ref{tab:diff}.

\begin{algorithm}[tb]
   \caption{$\mathcal{I}$-Evidential Deep Learning}
   \label{alg_train}
\begin{algorithmic}
   \STATE {\bfseries Input:} $\lambda$, Training set $\mathcal{D}  = \left\{(\vx_{i}, \vy_{i})\right\}_{i=1}^{N}$, batch size $b$, learning rate $\beta$, total epochs $T$
   \STATE Initialize $\vtheta$
   \FOR{$t=0,1,\cdots, T$}
   \STATE $\lambda_t = \min(1.0, t/T)$
       \FOR{$\mathcal{D}_b \sim \mathcal{D}$} 
           \FOR{$(\vx_i, \vy_i) \sim \mathcal{D}_b$}
           \STATE $\boldsymbol{\alpha}_i = f_{\vtheta}(\vx_i) + 1$
           \STATE $\hat{\boldsymbol{\alpha}}_i = \boldsymbol{\alpha}_i \odot(1 - \vy_i) + \vy_i$
           \STATE $\mathcal{L}_{i} = \mathcal{L}_{i}^{\mathcal{I}\text{-MSE}} - \lambda \mathcal{L}_{i}^{|\mathcal{I}|} + \lambda_t \mathcal{L}_{i}^{\text{KL}} $ \quad //see Eq.(\ref{obj_final})
           \ENDFOR
        \STATE $\vtheta \leftarrow \vtheta - \beta \nabla_\vtheta \mathcal{L}$ with $\mathcal{L} = \frac{1}{b} \sum_{i=1}^{b} \mathcal{L}_{i}$ 
        \ENDFOR
    \ENDFOR
\end{algorithmic}
\end{algorithm}

\begin{table}[t]
	\caption{Given a sample $(\vx_i, \vy_i)$, the difference in loss function between $\mathcal{I}$-EDL and EDL are marked in \textcolor{blue}{blue}.}
	\label{tab:diff}
	\begin{center}
        \renewcommand{\arraystretch}{2}
	\resizebox{0.48\textwidth}{!}{
		\begin{tabular}{c|cc}
			\toprule
			 &\textbf{EDL}   &\textbf{$\mathcal{I}$-EDL} \\ \midrule
              MSE  & \begin{tabular} {@{}c@{}} $\sum_{j=1}^{K} (y_{i j} - \frac{\alpha_{i j}}{\alpha_{i 0}} )^2$ \\ $+ \sum_{j=1}^{K} \frac{\alpha_{i j}(\alpha_{i 0} - \alpha_{i j})}{\alpha_{i 0}^2 (\alpha_{i 0} + 1)} $ \end{tabular} & \begin{tabular}{@{}c@{}} $\sum_{j=1}^{K} (y_{i j} - \frac{\alpha_{i j}}{\alpha_{i 0}} )^2\textcolor{blue}{\psi^{(1)}(\alpha_{i j})}$ \\ $+ \sum_{j=1}^{K} \frac{\alpha_{i j}(\alpha_{i 0} - \alpha_{i j})}{\alpha_{i 0}^2 (\alpha_{i 0} + 1)} \textcolor{blue}{\psi^{(1)}(\alpha_{i j})}$ \end{tabular} \\
              \hline
             KL & $\KL(Dir(\hat{\boldsymbol{\alpha}_i}) \Vert Dir(\boldsymbol{1}))$ & $\KL(Dir(\hat{\boldsymbol{\alpha}_i}) \Vert Dir(\boldsymbol{1}))$  \\
             \hline
             $\mathcal{I}$  & - & \textcolor{blue}{$- \log |\mathcal{I}(\boldsymbol{\alpha}_i)|$ }\\
			\bottomrule
		\end{tabular}}
	\end{center}
\end{table}
\section{Related Work}
\label{sec_related}

\textbf{Dirichlet-based uncertainty models (DBU)} predict the parameters of the Dirichlet distribution, which allows the computation of closed-form classical uncertainty metrics such as differential entropy, mutual information, etc. These metrics can be used to distinguish among data, model, and distributional uncertainty. Different DBU models differ in the parameterization and training strategy of the Dirichlet distribution. For example, \textbf{KL-PN} \cite{malinin2018predictive} proposes the Prior Networks (PN) trained with two KL divergence terms. The first term is used to learn sharp Dirichlet parameters for ID data, while the other learns flat Dirichlet parameters for OOD data. Since the forward KL divergence is zero-avoiding, \textbf{RKL-PN} \cite{malinin2019reverse} introduces the reverse KL divergence to avoid undesired multimodal target distributions. Posterior Network (\textbf{PostN}) \cite{charpentier2020posterior} uses Normalizing Flows to predict the posterior distribution of any input sample without training with OOD data. Evidential Deep Learning (\textbf{EDL}) \cite{sensoy2018evidential} treats the network's outputs as belief masses based on the Dempster-Shafer Theory of Evidence (DST) \cite{sentz2002combination} and derives the loss function using subjective logic \cite{josang2016subjective}. Moreover, deep evidential regression \cite{amini2020deep, soleimany2021evidential} introduces evidential priors over the original Gaussian likelihood function to model the uncertainty of regression networks. Meinert et al. \yrcite{meinert2022unreasonable} further analyze why DER can produce reasonable results in practice despite overparameterized representations of uncertainty. Zhao et al. \yrcite{zhao2020uncertainty} propose a multi-source uncertainty framework combined with DST for semi-supervised node classification with GNNs. Bao et al. \yrcite{bao2022opental} propose a general framework for Open Set Temporal Action Localization (OSTAL) based on EDL. Although EDL shows an impressive performance in uncertainty quantification and is widely used in various applications, recently proposed work \cite{bengs2022pitfalls} argues that classical EDL does not motivate learners to faithfully predict their epistemic uncertainty because it is sensitive to the regularization parameter. Compared with previous efforts, our method is the first to exploit evidence for training to improve the performance of uncertainty quantification. The Fisher information matrix (FIM) we introduce can be seen as some type of first-order distribution information, which can help learners make more accurate predictions and better estimate uncertainty.

\textbf{Bayesian Neural Networks (BNNs)} explicitly model network parameters as random variables, quantifying uncertainty by learning a posterior over parameters. Since the posterior inference of BNNs is intractable, many posterior approximation schemes have been developed to improve scalabilities, such as variational inference (VI) \cite{graves2011practical, blundell2015weight}, stochastic gradient Markov Chain Monte Carlo \cite{welling2011bayesian, ma2015complete}, and Laplace approximation \cite{ritter2018scalable, kristiadi2021learnable}. Furthermore, 
the integral of marginalizing the likelihood with the posterior distribution is also intractable and is typically approximated via sampling. A well-known method is Monte Carlo Dropout (\textbf{MC Dropout}) \cite{gal2016dropout}, which treats the dropout layer as a Bernoulli distributed random variable, and training a network with dropout layers can be interpreted as an approximate VI. However, these methods require significant modifications to the training process and are computationally expensive, and more importantly, cannot distinguish between distributional uncertainty and other uncertainties.

\textbf{Calibration methods} aim to reduce over-confidence by calibrating models. For example, Guo et al. \yrcite{guo2017calibration} introduce temperature scaling as a post-hoc calibration to mitigate overconfidence. ODIN \cite{liang2018enhancing} uses a mix of temperature scaling at the softmax layer and input perturbations. Pereyra et al. \yrcite{pereyra2017regularizing} penalize the low-entropy output distribution in the loss function. Karandikar et al. \yrcite{karandikar2021soft} propose differentiable losses to improve calibration based on a soft version of the binning operation underlying popular calibration-error estimators. Roelofs et al. \yrcite{roelofs2022mitigating} focus on assessing statistical bias in calibration. Since these methods cannot distinguish between different types of uncertainty, they are often combined with the first two types of methods, such as Lakshminarayanan et al. \yrcite{lakshminarayanan2017simple} combined calibration results with deep ensembles. 
Additionally, there are also some methods of modeling with label noise. For example, Collier et al. \yrcite{collier2021correlated} propose input-dependent noise losses for label noise in classification. Cui et al. \yrcite{cui2022confidence} provides a unified framework for reliable learning under the joint (image, label)-noise. Compared to these works, we focus on the evidence underestimation problem in evidential networks, and more importantly, we are the first to address this problem by introducing the Fisher information matrix.

\section{Experiments}
\label{tl_exp}

In this section, we conduct extensive experiments to compare the performance of our proposed method with previous methods on multiple uncertainty estimation-related tasks. See Appendix \ref{app_exp} for additional results and more details. \footnote{The code is available at: \url{https://github.com/danruod/IEDL}}

\subsection{Experimental Setup}
\label{exp_setup}

\paragraph{Datasets} 
We evaluate our algorithm on the following image classification datasets: \textbf{MNIST} \cite{lecun1998mnist}, \textbf{CIFAR10} \cite{krizhevsky2009learning}, and \textbf{mini-ImageNet} \cite{vinyals2016matching}. For OOD detection experiments, we use \textbf{KMNIST} \cite{clanuwat2018deep} and \textbf{FashionMNIST} \cite{xiao2017fashion} for MNIST, the Street View House Numbers (\textbf{SVHN}) \cite{netzer2018street} and \textbf{CIFAR100} 
 \cite{krizhevsky2009learning} for CIFAR10, and the Caltech-UCSD Birds (\textbf{CUB}) dataset \cite{wah2011caltech} for mini-ImageNet. More details are given in Appendix \ref{app_exp_dataset}.

\paragraph{Implementation details}
Following Charpentier et al. \yrcite{charpentier2020posterior}, we use 3 convolutional layers and 3 dense layers (ConvNet) on MNIST and VGG16 \cite{simonyan2014very} on CIFAR10. For all experiments on both datasets, we split the data into train, validation, and test sets. We use a validation loss-based early termination strategy to train up to $200$ epochs with a batch size of $64$. For the mini-ImageNet dataset, we conduct experiments on the more challenging few-shot classification setting. We use WideResNet-28-10 pre-trained backbone from Yang et al. \yrcite{yang2021free} as the feature extractor to train a 1-layer classifier. Refer to Appendix \ref{app_exp_implem} for more details.

\begin{table*}[t]
	\caption{AUPR scores of OOD detection (mean $\pm$ standard deviation of 5 runs). $^\dag$ indicates that the first four lines are the results reported by Charpentier et al. \yrcite{charpentier2020posterior}. Bold and underlined numbers indicate the best and runner-up scores, respectively. }
	\label{tab_ood}
	\begin{center}
	\resizebox{\textwidth}{!}{
            \begin{tabular}{ccc|cc|cc|cc}
            \toprule
			&\multicolumn{2}{c}{\textbf{MNIST} $\rightarrow$ \textbf{KMNIST}$^\dag$}  &\multicolumn{2}{c}{\textbf{MNIST} $\rightarrow$ \textbf{FMNIST}$^\dag$} &\multicolumn{2}{c}{\textbf{CIFAR10} $\rightarrow$ \textbf{SVHN}$^\dag$}  &\multicolumn{2}{c}{\textbf{CIFAR10} $\rightarrow$ \textbf{CIFAR100}}  \\ 
             \cmidrule(lr){2-3}  \cmidrule(lr){4-5} \cmidrule(lr){6-7} \cmidrule(lr){8-9} 
			\textbf{Method}  &\textbf{Max.P}   &\textbf{$\alpha_0$} &\textbf{Max.P}   &\textbf{$\alpha_0$} &\textbf{Max.P}   &\textbf{$\alpha_0$} &\textbf{Max.P}   &\textbf{$\alpha_0$} \\ 
            \midrule
			MC Dropout & 94.00 $\pm$ 0.1 & - & 96.56 $\pm$ 0.2 & - & 51.39 $\pm$ 0.1 & - & 45.57 $\pm$ 1.0 & - \\
			KL-PN &92.97 $\pm$ 1.2 & 93.39 $\pm$ 1.0 & \underline{98.44 $\pm$ 0.1} & \underline{98.16 $\pm$ 0.0} & 43.96 $\pm$ 1.9 & 43.23 $\pm$ 2.3 & 61.41 $\pm$ 2.8 & 61.53 $\pm$ 3.4 \\
                RKL-PN & 60.76 $\pm$ 2.9 & 53.76 $\pm$ 3.4 &78.45 $\pm$ 3.1 & 72.18 $\pm$ 3.6 & 53.61 $\pm$ 1.1 & 49.37 $\pm$ 0.8 & 55.42 $\pm$ 2.6 & 54.74 $\pm$ 2.8 \\
                PostN & 95.75 $\pm$ 0.2 & 94.59 $\pm$ 0.3 & 97.78 $\pm$ 0.2 & 97.24 $\pm$ 0.3 & \underline{80.21 $\pm$ 0.2} & 77.71 $\pm$ 0.3 & 81.96 $\pm$ 0.8 & 82.06 $\pm$ 0.8 \\
                EDL & 97.02 $\pm$ 0.8 & \underline{96.31 $\pm$ 2.0} & 98.10 $\pm$ 0.4 & 98.08 $\pm$ 0.4 & 78.87 $\pm$ 3.5 & \underline{79.12 $\pm$ 3.7} & \underline{84.30 $\pm$ 0.7} & \underline{84.18 $\pm$ 0.7} \\
                \midrule
                \textbf{$\mathcal{I}$-EDL} & \textbf{98.34 $\pm$ 0.2} & \textbf{98.33 $\pm$ 0.2} & \textbf{98.89 $\pm$ 0.3} & \textbf{98.86 $\pm$ 0.3} & \textbf{83.26 $\pm$ 2.4} & \textbf{82.96 $\pm$ 2.2} & \textbf{85.35 $\pm$ 0.7} & \textbf{84.84 $\pm$ 0.6} \\
            \bottomrule
        \end{tabular}}
    \end{center}
\end{table*}

\paragraph{Baselines}
We focus on comparing our algorithm with other Dirichlet-based uncertainty methods, since only DBU methods can distinguish different types of uncertainty compared to BNNs and calibration methods, as mentioned previously. In particular, we compare to following baselines: Prior Networks (PN) trained with KL divergence (\textbf{KL-PN}) \cite{malinin2018predictive} and Reverse KL divergence (\textbf{RKL-PN}) \cite{malinin2019reverse}, Posterior Network (\textbf{PostN}) \cite{charpentier2020posterior}, and Evidential Deep Learning (\textbf{EDL}) \cite{sensoy2018evidential}. 
Furthermore, we compare the dropout model (\textbf{MC Dropout}) \cite{gal2016dropout}, which is often state-of-the-art in many uncertainty estimation tasks \cite{ovadia2019can}. Since other methods except KL-PN and RKL-PN do not require OOD data for training. For a fair comparison, we use the uniform noise instead of actual OOD test data as OOD training data for the former two methods following Charpentier et al. \yrcite{charpentier2020posterior}.

\begin{table}[t]
\vspace{-0.2cm}
	\caption{AUPR scores and accuracy of CIFAR10 with VGG16 in misclassified image detection and image classification, respectively. Each experiment is run with 5 seeds. $^\dag$ denotes results reported by \cite{charpentier2020posterior}. Bold and underlined numbers indicate the best and runner-up scores, respectively.}
	\label{tab_conf}
	\begin{center}
	\resizebox{0.47\textwidth}{!}{
		\begin{tabular}{c|cc|c}
		\toprule
			\textbf{Method}      &\textbf{Max.P} &\textbf{Max.$\alpha$}   &\textbf{Acc.} \\ 
            \midrule
			MC Dropout$^\dag$ & 97.15 $\pm$ 0.0 & - & 82.84 $\pm$ 0.1  \\
                KL-PN$^\dag$ & 50.61 $\pm$ 4.0 & 52.49 $\pm$ 4.2 & 27.46 $\pm$ 1.7  \\
                RKL-PN$^\dag$ & 86.11 $\pm$ 0.4 & 85.59 $\pm$ 0.3 & 64.76 $\pm$ 0.3 \\
                PostN$^\dag$ & 97.76 $\pm$ 0.0 & 97.25 $\pm$ 0.0 & \underline{84.85 $\pm$ 0.0} \\
                EDL & \underline{97.86 $\pm$ 0.2} & \underline{97.86 $\pm$ 0.2} & 83.55 $\pm$ 0.6 \\
                \midrule
                \textbf{$\mathcal{I}$-EDL} & \textbf{98.72 $\pm$ 0.1} & \textbf{98.63 $\pm$ 0.1} & \textbf{89.20 $\pm$ 0.3} \\
            \bottomrule
        \end{tabular}}
    \end{center}
    \vspace{-0.3cm}
\end{table}

\subsection{Confidence Evaluation}
\label{exp_conf}

We first measure the availability of uncertainty estimates in the confidence evaluation tasks that aim to answer an interesting question "\textit{Are more confident (i.e., less uncertain) predictions more likely to be correct?}". We use the area under the precision-recall curve (\textbf{AUPR}) metric. For DBU methods, we represent $\max_{c} p_c$ (Max.P) and $\max_{c} \alpha_c$ (Max.$\alpha$) respectively as the scores with labels $1$ for correct and $0$ for incorrect predictions. Since the dropout model does not have concentration parameters, we only provide results with Max.P as scores.
Table \ref{tab_conf} shows our proposed method achieves state-of-the-art performance in all measurements. In particular, our method improves image classification by about $\textbf{5.2\%}$ and confidence estimation by about $\textbf{0.9\%}$ compared to the runner-up methods.

\subsection{OOD detection}
\label{exp_ood}

We then measure the usability of uncertainty quantification in the OOD detection task. The performance of OOD detection is also measured by \textbf{AUPR} with labels $1$ for ID data and $0$ for OOD data. 
The scores of the DBU methods are given by $\max_{c} p_c$ (Max.P) and $\alpha_0$ ($\sum_{c} \alpha_c$) respectively, while Dropout uses Max.P as scores. We compare our method with other methods on four OOD detection tasks, including MNIST against KMNIST and FMNIST, and CIFAR10 against SVHN and CIFAR100. Table \ref{tab_ood} shows our proposed method achieves superior performance in all tasks without training with additional OOD data. More specifically, $\mathcal{I}$-EDL outperforms the second-placed method by about $\textbf{1.3\%}$, $\textbf{0.5\%}$, $\textbf{3.8\%}$ and $\textbf{1.2\%}$ on four OOD detection tasks, respectively. Note that EDL does not achieve suboptimal performance on all OOD detection tasks. We also evaluate performance using differential entropy (\textbf{D.Ent.}) and mutual information (\textbf{M.I.}) as scores and area under a ROC curve (\textbf{AUROC}). All these results can be seen in Appendix \ref{app_exp_results}. Given lots of efforts contributed to OOD detection \cite{liang2018enhancing,sastry2020detecting}, here we mainly focus on the comparisons with DBU models, which solve OOD detection by distinguishing different types of uncertainty.

\subsection{Few-shot Learning}
\label{exp_fsl}

\begin{table*}[t]
    \caption{Classification accuracy (Acc.), AUPR scores for both confidence evaluation (Conf.) and OOD detection (OOD) under $\{5, 10\}$-way $\{1, 5, 20\}$-shot settings of mini-ImageNet. CUB is used for OOD detection. Each experiment is run for over $10,000$ few-shot episodes.}
	\label{tab_mini}
	\begin{center}
	\resizebox{\textwidth}{!}{
		\begin{tabular}{cccc|ccc|ccc}
		\toprule
			&\multicolumn{3}{c}{\textbf{5-Way 1-Shot}}  &\multicolumn{3}{c}{\textbf{5-Way 5-Shot}} &\multicolumn{3}{c}{\textbf{5-Way 20-Shot}}  \\ 
             \cmidrule(lr){2-4}  \cmidrule(lr){5-7} \cmidrule(lr){8-10}
             \textbf{Method}     &\textbf{Acc.}  &\textbf{Conf. (Max.$\alpha$)}   &\textbf{OOD ($\alpha_0$)}  &\textbf{Acc.}  &\textbf{Conf. (Max.$\alpha$)}   &\textbf{OOD ($\alpha_0$)} &\textbf{Acc.}  &\textbf{Conf. (Max.$\alpha$)}   &\textbf{OOD ($\alpha_0$)} \\ \midrule
			EDL & 61.00 $\pm$ 0.22 & 80.59 $\pm$ 0.23 & 65.40 $\pm$ 0.26 & 80.38 $\pm$ 0.15 & 93.92 $\pm$ 0.09 & 76.53 $\pm$ 0.27 & 85.54 $\pm$ 0.12 & 97.51 $\pm$ 0.04 & 79.78 $\pm$ 0.23 \\
            \textbf{$\mathcal{I}$-EDL} & 63.82 $\pm$ 0.20 & 82.00 $\pm$ 0.21 & 74.76 $\pm$ 0.25 & 82.00 $\pm$ 0.14 & 94.09 $\pm$ 0.09 & 82.48 $\pm$ 0.20 & 88.12 $\pm$ 0.09 & 97.54 $\pm$ 0.04 & 85.40 $\pm$ 0.19 \\
            \midrule
            \textbf{$\Delta$} &\textbf{2.82} & \textbf{1.41} & \textbf{9.36} & \textbf{1.62} & \textbf{0.17} & \textbf{5.95} & \textbf{2.58} & \textbf{0.04} & \textbf{5.62} \\
                \midrule
            \midrule
                &\multicolumn{3}{c}{\textbf{10-Way 1-Shot}}  &\multicolumn{3}{c}{\textbf{10-Way 5-Shot}} &\multicolumn{3}{c}{\textbf{10-Way 20-Shot}}  \\ 
             \cmidrule(lr){2-4}  \cmidrule(lr){5-7} \cmidrule(lr){8-10}
             \textbf{Method}     &\textbf{Acc.}  &\textbf{Conf. (Max.$\alpha$)}   &\textbf{OOD ($\alpha_0$)}  &\textbf{Acc.}  &\textbf{Conf. (Max.$\alpha$)}   &\textbf{OOD ($\alpha_0$)} &\textbf{Acc.}  &\textbf{Conf. (Max.$\alpha$)}   &\textbf{OOD ($\alpha_0$)} \\ \midrule
			EDL & 44.55 $\pm$ 0.15 & 65.97 $\pm$ 0.20 & 67.83 $\pm$ 0.24 & 62.52 $\pm$ 0.16 & 86.81 $\pm$ 0.10 & 76.34 $\pm$ 0.20 & 69.29 $\pm$ 0.17 & 94.21 $\pm$ 0.06 & 76.88 $\pm$ 0.17 \\
			\textbf{$\mathcal{I}$-EDL} &49.37 $\pm$ 0.13 & 68.29 $\pm$ 0.19 & 71.95 $\pm$ 0.20 & 67.89 $\pm$ 0.11 & 87.45 $\pm$ 0.09 & 82.29 $\pm$ 0.19 & 78.60 $\pm$ 0.08 & 94.40 $\pm$ 0.04 & 82.52 $\pm$ 0.14 \\
            \midrule
            \textbf{$\Delta$} &\textbf{4.82} & \textbf{2.32} & \textbf{4.12} & \textbf{5.37} & \textbf{0.64} & \textbf{5.95} & \textbf{9.31} & \textbf{0.19} & \textbf{5.64} \\
            \bottomrule
        \end{tabular}}
    \end{center}
\end{table*}

We next conduct more challenging few-shot experiments on mini-ImageNet. We use the WideResNet trained following Yang et al. \yrcite{yang2021free} to obtain pre-trained features, and then train the 1-layer classifier under $N$-way $K$-shot setting. We evaluate $\{5, 10\}$-way $\{1, 5, 20\}$-shot classification, confidence estimation and OOD detection. The performance of classification and uncertainty estimation are reported in the average accuracy($\%$,  top-1) and AUPR($\%$), respectively, with $95\%$ confidence interval over $10,000$ few-shot episodes. Each episode contains randomly sampled $N$ classes and $K$ samples per class for adaptation, $\min(15, K)$ query samples per class for image classification and confidence evaluation, and the same number of query samples from CUB for OOD detection.

As shown in Table \ref{tab_mini}, the average test accuracy, Max.$\alpha$-based confidence evaluation, and $\alpha_0$-based OOD detection show impressive improvements on $\mathcal{I}$-EDL over EDL on all the $N$-way $K$-shot tasks. More specifically, all average test accuracy improvements of our method exceed $\textbf{1.62\%}$, up to $\textbf{9.31\%}$ under $10$-way $20$-shot. In confidence evaluation, $\mathcal{I}$-EDL also shows better performance than EDL, especially the improvement over $\textbf{2.32\%}$ under $10$-way $1$-shot. Moreover, $\mathcal{I}$-EDL shows excellent performance in OOD detection, where all improvements are between $\textbf{4.12\%}$ and $\textbf{9.36\%}$. Furthermore, we compare the OOD detection performance of our method and other DBU methods under $5$-way $5$-shot in Fig. \ref{fig_fsl_noisy}(a).
All of these results demonstrate that our method not only improves classification accuracy but also greatly improves the availability of uncertainty estimation in the more challenging few-shot scenarios. More results including AUPR and AUROC scores based on M.I. and D.Ent are provided in Appendix \ref{app_exp_results_fsl}.

\subsection{Noisy data detection}
\label{exp_noisy}

\begin{figure}[t]
    \centering
    \includegraphics[width=0.48\textwidth]{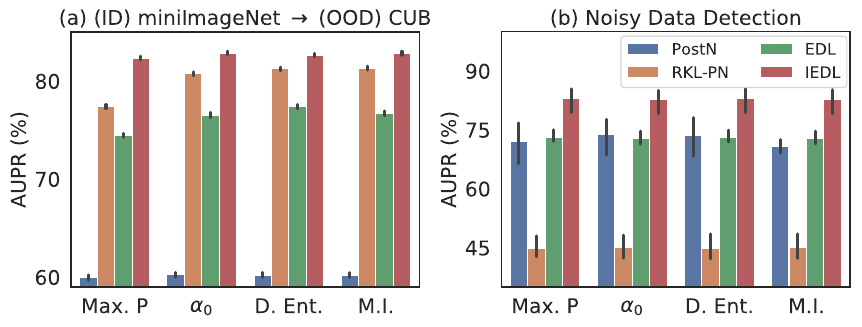}
    \vspace{-3mm}
    \caption{(a) Comparison of OOD detection (AUPR) on mini-ImageNet against CUB under $5$-way $5$-shot. (b) Noisy data detection on CIFAR10.}
    \label{fig_fsl_noisy}
\end{figure}

We finally evaluate our method on noisy examples. Noisy examples are generated by adding zero-mean isotropic Gaussian noise with standard deviation $\sigma = 0.1$ to the test data of the ID dataset. As shown in Fig. \ref{fig_fsl_noisy}(b), our method demonstrates excellent detection of noisy data across all metrics, outperforming the runner-up method by more than $11\%$.

\subsection{Ablation Study}
\label{exp_ablation}

\begin{table}[t]
\vspace{-3mm}
	\caption{Ablation studies under mini-ImageNet 5-way 5-shot for image classification, confidence evaluation, and OOD detection against CUB.}
	\label{tab_ablation}
	\begin{center}
	\resizebox{0.48\textwidth}{!}{
		\begin{tabular}{cc|ccc}
			\toprule
			\textbf{$\mathcal{I}$-MSE} &$|\mathcal{I}|$  &\textbf{Acc.} &\textbf{Conf. (Max.$\alpha$)}   &\textbf{OOD ($\alpha_0$)}  \\ \midrule
                &  & 80.38 $\pm$ 0.15 & 93.92 $\pm$ 0.09 & 76.53 $\pm$ 0.27 \\
                &\checkmark  & 81.82 $\pm$ 0.14  & 93.97 $\pm$ 0.09 & 79.68 $\pm$ 0.25 \\
			\checkmark & & 81.27 $\pm$ 0.14 & \textbf{94.42} $\pm$ 0.08 & 81.75 $\pm$ 0.22 \\
			\checkmark   &\checkmark  & \textbf{82.00} $\pm$ 0.14 & 94.09 $\pm$ 0.09 &\textbf{82.48} $\pm$ 0.20 \\
			\bottomrule
		\end{tabular}}
	\end{center}
\vspace{-3mm}
\end{table}

We further investigate our method performance with an ablation study and summarize it in Table \ref{tab_ablation}. We respectively ablate the effects of expected FIM-weighted MSE ($\mathcal{I}$-MSE) and FIM's negative log determinant ($|\mathcal{I}|$). Note that classical EDL is equivalent to $\mathcal{I}$-EDL without using $\mathcal{I}$-MSE and $|\mathcal{I}|$. From the result, we observe that both optimizations are beneficial for image classification, confidence evaluation, and OOD detection. In particular, with the only usage of $|\mathcal{I}|$ ($\mathcal{I}$-MSE), the improvements over EDL for classification and $\alpha_0$-based OOD detection are $\sim \textbf{1.9\%}$ ($\textbf{1.3\%}$) and $\sim \textbf{4.1\%}$ ($\textbf{6.8\%}$). If both optimizations are used together, the improvements increase to about $\textbf{2.0\%}$ and $\textbf{7.8\%}$. Thus, the combination of two optimizations achieves a win-win effect. More ablation studies are provided in Appendix \ref{app_exp_hyper}.

\subsection{Analysis of Uncertainty estimation}
\label{exp_unc}
\begin{figure}[t]
    \centering
    \includegraphics[width=0.48\textwidth]{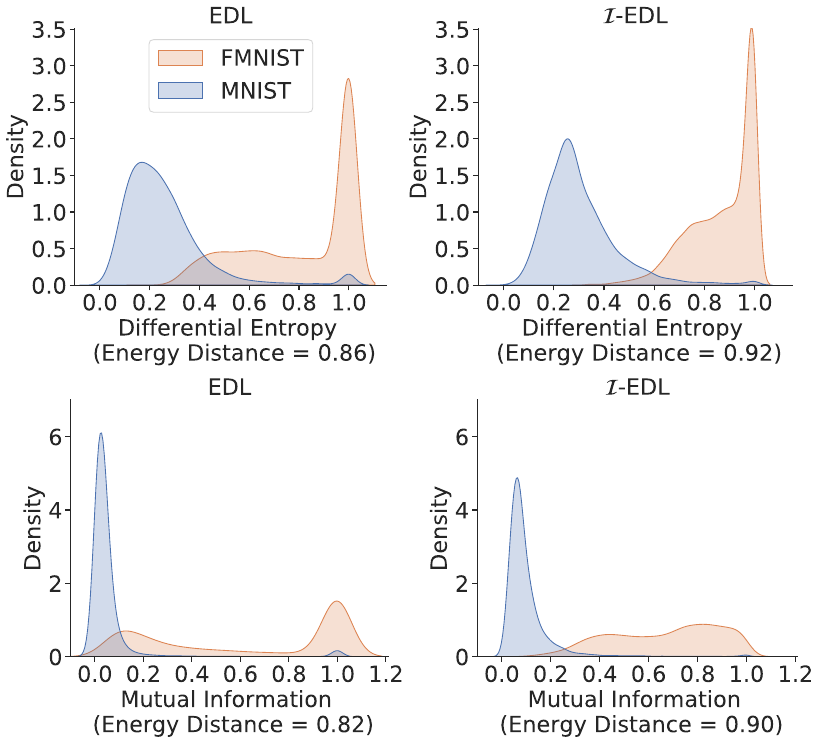}
    \vspace{-5mm}
    \caption{Uncertainty representation for ID (MNIST) and OOD (FMNIST). More results are shown in Appendix \ref{app_exp_unc}.}
    \label{fig_density_mnist_diff_ent}
\end{figure}

Figure \ref{fig_density_mnist_diff_ent} represents density plots of the predicted differential entropy and mutual information\footnote{Detailed formulas refer to Eq.(\ref{eq_diff_ent}) and Eq.(\ref{eq_mi}) in Appendix \ref{app_unc}}. Lower entropy or mutual information represents the model yields a sharper distribution, indicating that the sample has low uncertainty. We also report the energy distance \cite{szekely2013energy} of two distributions (Formula is given in Appendix \ref{app_energy}.), which shows that our method provides more separable uncertainty estimates. More specifically, $\mathcal{I}$-EDL produces sharper prediction peaks than EDL, both in the low uncertainty region of ID samples and the high uncertainty region of OOD samples. Furthermore, our method also reduces the occurrence of ID samples in high-uncertainty regions.

\section{Conclusion}
\label{sec_conclu}
In this paper, we found that the classical EDL trained with mean square error would hinder the learning of evidence, especially for high data uncertainty samples. To address this issue, we propose a novel and simple method, \textit{Fisher Information-based Evidential Deep Learning} ($\mathcal{I}$-EDL) to alleviate the over-penalization of the mislabeled classes by considering importance weights with different classes for each sample. More specifically, we introduce the perspective of generative models to model evidential networks, where the observed label is jointly generated by the predicted categorical probability and the informativeness of each class contained in the sample. The categorical probabilities are generated from the Dirichlet distribution with its concentration parameter calculated by passing the input sample through the evidential network, while the information is obtained from FIM. Extensive experiments on various image classification, confidence evaluation and OOD detection tasks, as well as comparisons with some state-of-the-art algorithms, demonstrate the effectiveness of our approach in achieving high classification and uncertainty quantification.
Since our method is designed based on Dirichlet distribution, it cannot be directly applied to regression tasks. A naive approach is to discretize the regression task, but this is not a good solution because it loses information about continuous labels, such as order. However, we believe that avoiding overfitting caused by data uncertainty in regression tasks and how to apply our ideas to regression tasks is a very attractive problem worthy of exploration in future work.

\section*{Acknowledgments}
\label{sec_ack} 
We would like to thank Prof. Chang-Yu Hsieh for his helpful discussions on Evidential Deep Learning, and anonymous reviewers for their feedback and constructive comments to improve this work. This work is supported by the National Key R\&D Program of China (2022YFE0200700), the Research Grants Council of the Hong Kong Special Administrative Region, China (Project Reference Number: T45-401/22-N and 14201321), the National Natural Science Foundation of China (Project No. 62006219), the Natural Science Foundation of Guangdong Province (2022A1515011579), and the Hong Kong Innovation and Technology Fund (Project No. ITS/170/20 and ITS/241/21).




\bibliography{main}
\bibliographystyle{icml2023}

\newpage
\appendix
\onecolumn
\section{Derivation and Proof}

This section provides the derivation of the Fisher Information Matrix (FIM) of Dirichlet distribution and the final objective function of Eq. \ref{obj_final}. We also provide a brief overview of the proof of Theorem \ref{pac} from \cite{germain2009pac, alquier2016properties, masegosa2020learning}.

\subsection{FIM Derivation for Dirichlet Distribution}
\label{app_proof_fisher}

The Fisher Information Matrix (FIM) of Dirichlet distribution is defined as:
$$
\mathcal{I}(\boldsymbol{\alpha}) = \E \left[ \frac{\partial \ell}{\partial \boldsymbol{\alpha}} \frac{\partial \ell}{\partial \boldsymbol{\alpha}^{T}}\right] \in \mathbb{R}^{K \times K},
$$
where $\ell = \log Dir(\vp|\boldsymbol{\alpha})$ is the log-likelihood function. Under weak conditions (see Lemma 5.3 in \cite{lehmann2006theory}), the FIM can be expressed as $\mathcal{I}(\boldsymbol{\alpha}) = \E_{Dir(\vp|\boldsymbol{\alpha})} \left[ - \partial^2 \ell /  \partial \boldsymbol{\alpha} \boldsymbol{\alpha}^{T} \right]$. Thus, we can calculate each element by
$$
\begin{aligned}
 \relax[\mathcal{I}(\boldsymbol{\alpha})]_{i j} &= \E_{Dir(\vp|\boldsymbol{\alpha})} \left[ - \frac{\partial^2}{\partial \alpha_i \partial \alpha_j } \log Dir(\vp|\boldsymbol{\alpha}) \lvert \boldsymbol{\alpha} \right] \\
& = \E_{Dir(\vp|\boldsymbol{\alpha})} \left[ - \frac{\partial^2}{\partial \alpha_i \partial \alpha_j } \left( \log \Gamma\left(\alpha_{0}\right) - \sum_{k=1}^{K} \log \Gamma\left(\alpha_{k}\right) + \sum_{k=1}^{K} (\alpha_{k} - 1) \log p_{k} \right) \right] \\
& = \E_{Dir(\vp|\boldsymbol{\alpha})} \left[ - \frac{\partial}{\partial \alpha_j } \left( \psi\left(\alpha_{0}\right) - \psi\left(\alpha_{i}\right) + \log p_{i} \right) \right] \\
& = \begin{cases}
\psi^{(1)}\left(\alpha_{i}\right) - \psi^{(1)}\left(\alpha_{0}\right), & i=j, \\ 
-\psi^{(1)}\left(\alpha_{0}\right), & i \neq j,
\end{cases}
\end{aligned}
$$
where $\Gamma(\cdot)$ is the \textit{gamma} function, $\psi(\cdot)$ is the \textit{digamma} function, $\psi^{(1)}(\cdot)$ is the \textit{trigamma} function, defined as $\psi^{(1)}(x) = d \psi(x) / dx = d^2 \ln \Gamma(x) / dx^2$. Then, we can get the matrix form of the FIM:
\begin{equation}
\begin{aligned}
\label{fisher_matrix}
\mathcal{I}(\boldsymbol{\alpha}) = &
\begin{bmatrix}
\psi^{(1)}\left(\alpha_{1}\right) - \psi^{(1)}\left(\alpha_{0}\right) & - \psi^{(1)}\left(\alpha_{0}\right) & \cdots & - \psi^{(1)}\left(\alpha_{0}\right) \\
- \psi^{(1)}\left(\alpha_{0}\right) & \psi^{(1)}\left(\alpha_{2}\right) - \psi^{(1)}\left(\alpha_{0}\right) & \cdots & - \psi^{(1)}\left(\alpha_{0}\right) \\
\vdots & \vdots & \ddots & \vdots \\
- \psi^{(1)}\left(\alpha_{0}\right) & - \psi^{(1)}\left(\alpha_{0}\right) & \cdots & \psi^{(1)}\left(\alpha_{K}\right) - \psi^{(1)}\left(\alpha_{0}\right)
\end{bmatrix} \\
= & \ \text{diag}([\psi^{(1)}(\alpha_{1}), \cdots, \psi^{(1)}(\alpha_{K})])- \psi^{(1)}(\alpha_{0}) \boldsymbol{1}\boldsymbol{1}^T,
\end{aligned}
\end{equation}
where $\boldsymbol{1} = [1; \cdots; 1] \in \mathbb{R}^{K}$.
Let $\vb=[ \psi^{(1)}(\alpha_{1}), \cdots, \psi^{(1)}(\alpha_{K})]^T$, by applying Matrix-Determinant Lemma, we have
$$
|\mathcal{I}(\boldsymbol{\alpha})| = |\text{diag}(\vb)| \cdot (1 - \psi^{(1)}(\alpha_{0}) \boldsymbol{1}^T \text{diag}(\vb)^{-1}\boldsymbol{1}) = \prod_{i=1}^{K} \psi^{(1)}(\alpha_{i}) (1 - \sum_{j=1}^{K} \frac{\psi^{(1)}(\alpha_{0})}{\psi^{(1)}(\alpha_{j})}).
$$
Therefore,
\begin{equation}
\label{log_det_fisher}
\log |\mathcal{I}(\boldsymbol{\alpha})| = \sum_{i=1}^{K} \log \psi^{(1)}(\alpha_{i}) + \log (1 - \sum_{i=1}^{K} \frac{\psi^{(1)}(\alpha_{0})}{\psi^{(1)}(\alpha_{i})}).
\end{equation}

\subsection{Derivation of the objective function Eq. \ref{obj_final}}
\label{app_proof_obj}

Given training set $\mathcal{D} = \{(\vx_i, \vy_i)\}_{i=1}^{N}$, by applying Theorem \ref{pac}, the upper bound of Eq.(\ref{obj_func_exp}) can be expressed as
\begin{equation}
\begin{aligned}
\label{app_obj_func_exp}
\min_{\vtheta} \quad &\E_{(\vx, \vy) \sim \mathcal{D}} \left[ \E_{\vp \sim Dir(\boldsymbol{\alpha})} \left[- \log p(\vy|\vp, \boldsymbol{\alpha}, \sigma^2)\right] + \lambda \KL(Dir(\vp | \hat{\boldsymbol{\alpha}}) \Vert Dir(\vp|\boldsymbol{1}) ) \right] \\
\text{s.t.} \quad &\boldsymbol{\alpha} = f_{\vtheta}(\vx) + 1 \\
&\hat{\boldsymbol{\alpha}} = \boldsymbol{\alpha} \odot(1 - \vy) + \vy \\
&\mathcal{I}(\boldsymbol{\alpha}) = \text{diag}([\psi^{(1)}(\alpha_{1}), \cdots, \psi^{(1)}(\alpha_{K})])- \psi^{(1)}(\alpha_{0}) \boldsymbol{1}\boldsymbol{1}^T \\
& \vp \sim Dir(\boldsymbol{\alpha}) \\
& p(\vy|\vp, \boldsymbol{\alpha}, \sigma^2) = \mathcal{N}(\vy|\vp, \sigma^2 \mathcal{I}(\boldsymbol{\alpha})^{-1}). \\
\end{aligned}
\end{equation}

We first simplify the first term $\E_{Dir(\vp|\boldsymbol{\alpha})} [ - \log \mathcal{N}(\vy|\vp, \sigma^2\mathcal{I}(\boldsymbol{\alpha})^{-1})]$ (abbreviated as $\E[ - \log \mathcal{N}(\vy|\vp, \sigma^2\mathcal{I}(\boldsymbol{\alpha})^{-1})]$),
\begin{equation}
\begin{aligned}
\label{app_objective_func}
& \E [ - \log \mathcal{N}(\vy|\vp, \sigma^2\mathcal{I}(\boldsymbol{\alpha})^{-1})] \\
= & \E \left[ - \log \left( (2\pi\sigma^2)^{-\frac{k}{2}} |\mathcal{I}(\boldsymbol{\alpha})|^{\frac{1}{2}} \exp ( -\frac{1}{2\sigma^2}(\vy - \vp)^T \mathcal{I}(\boldsymbol{\alpha})(\vy - \vp) ) \right) \right] \\
= & \E \left[ \frac{k}{2} \log (2\pi\sigma^2) - \frac{1}{2} \log |\mathcal{I}(\boldsymbol{\alpha})| + \frac{1}{2\sigma^2}(\vy - \vp)^T \mathcal{I}(\boldsymbol{\alpha})(\vy - \vp) \right] \\
= & \frac{k}{2} \log (2\pi\sigma^2) - \frac{1}{2} \log |\mathcal{I}(\boldsymbol{\alpha})| + \frac{1}{2\sigma^2}\E \left[ (\vy - \vp)^T \mathcal{I}(\boldsymbol{\alpha})(\vy - \vp)\right] 
\end{aligned}
\end{equation}

Then, $\E \left[ (\vy - \vp)^T \mathcal{I}(\boldsymbol{\alpha})(\vy - \vp)\right]$ can be converted to:
\begin{equation}
\begin{aligned}
\label{app_exp_mu_fisher}
& \E \left[ (\vy - \vp)^T \mathcal{I}(\boldsymbol{\alpha})(\vy - \vp)\right]  \\
= & \vy^T\mathcal{I}(\boldsymbol{\alpha})\vy - 2 \vy^T\mathcal{I}(\boldsymbol{\alpha})\E[\vp] + \E \left[ \vp^T \mathcal{I}(\boldsymbol{\alpha}) \vp\right] \\
= & \vy^T\mathcal{I}(\boldsymbol{\alpha})\vy - 2 \vy^T\mathcal{I}(\boldsymbol{\alpha})\E[\vp] +  \E[\vp]^T \mathcal{I}(\boldsymbol{\alpha}) \E[\vp] - \E[\vp]^T \mathcal{I}(\boldsymbol{\alpha}) \E[\vp] + \E \left[ \vp^T \mathcal{I}(\boldsymbol{\alpha}) \vp\right] \\
= &(\vy - \E[\vp] )^T \mathcal{I}(\boldsymbol{\alpha})(\vy - \E[\vp]) + \E \left[ (\vp - \E[\vp])^T \mathcal{I}(\boldsymbol{\alpha})(\vp - \E[\vp])\right] \\
\end{aligned}
\end{equation}


Since $\E[\vp] = \boldsymbol{\alpha} / \alpha_0$, $\Cov(p_i, p_j) = (\delta_{i j} \alpha_{i}\alpha_{0} - \alpha_{i}\alpha_{j}) / ( \alpha_{0}^2 (\alpha_{0} + 1))$, where $\delta_{i j}$ is the Kronecker delta (i.e. $\delta_{i j} = 1$ if $i=j$, else $\delta_{i j} = 0$), combined with the value of the FIM (Eq. \ref{fisher_matrix}), we have 

\begin{equation}
\begin{aligned}
\label{app_exp_mu_fisher_mse}
&(\vy - \E[\vp] )^T \mathcal{I}(\boldsymbol{\alpha})(\vy - \E[\vp]) \\
= &(\vy - \frac{\boldsymbol{\alpha}}{\alpha_0} )^T \begin{bmatrix}
\psi^{(1)}\left(\alpha_{1}\right) - \psi^{(1)}\left(\alpha_{0}\right) & - \psi^{(1)}\left(\alpha_{0}\right) & \cdots & - \psi^{(1)}\left(\alpha_{0}\right) \\
- \psi^{(1)}\left(\alpha_{0}\right) & \psi^{(1)}\left(\alpha_{2}\right) - \psi^{(1)}\left(\alpha_{0}\right) & \cdots & - \psi^{(1)}\left(\alpha_{0}\right) \\
\vdots & \vdots & \ddots & \vdots \\
- \psi^{(1)}\left(\alpha_{0}\right) & - \psi^{(1)}\left(\alpha_{0}\right) & \cdots & \psi^{(1)}\left(\alpha_{K}\right) - \psi^{(1)}\left(\alpha_{0}\right)
\end{bmatrix} (\vy - \frac{\boldsymbol{\alpha}}{\alpha_0} ) \\
= & (\vy - \frac{\boldsymbol{\alpha}}{\alpha_0} )^T \left(\text{diag}([\psi^{(1)}(\alpha_{1}), \cdots, \psi^{(1)}(\alpha_{K})])- \psi^{(1)}(\alpha_{0}) \boldsymbol{1}\boldsymbol{1}^T \right) (\vy - \frac{\boldsymbol{\alpha}}{\alpha_0} ) \\
= & \sum_{i=1}^{K} (y_i - \frac{\alpha_{i}}{\alpha_{0}} )^2 \psi^{(1)}\left(\alpha_{i}\right),
\end{aligned}
\end{equation}
and
\begin{equation}
\begin{aligned}
\label{app_exp_mu_fisher_var}
&\E \left[ (\vp - \E[\vp])^T \mathcal{I}(\boldsymbol{\alpha})(\vp - \E[\vp])\right] \\
= &\sum_{i, j=1}^{K} \Cov(p_i, p_j)\mathcal{I}(i, j) = \sum_{i, j =1}^{K} \frac{ \delta_{i j}\alpha_{i}\alpha_{0} - \alpha_{i}\alpha_{j}}{\alpha_{0}^2 (\alpha_{0} + 1)} (\delta_{i j} \psi^{(1)}\left(\alpha_{i}\right) - \psi^{(1)}\left(\alpha_{0}\right)) \\
= &\sum_{i=1}^{K} \frac{\alpha_{i}\alpha_{0} - \alpha_{i}^2}{\alpha_{0}^2 (\alpha_{0} + 1)} \psi^{(1)}\left(\alpha_{i}\right) - \sum_{i=1}^{K} \frac{\alpha_{i}\alpha_{0}}{\alpha_{0}^2 (\alpha_{0} + 1)}\psi^{(1)}\left(\alpha_{0}\right) + \sum_{i,j = 1}^{K} \frac{\alpha_{i}\alpha_{j}}{\alpha_{0}^2 (\alpha_{0} + 1)} \psi^{(1)}\left(\alpha_{0}\right) \\
= &\sum_{i=1}^{K} \frac{\alpha_{i}(\alpha_{0} - \alpha_{i})}{\alpha_{0}^2(\alpha_{0} + 1)} \psi^{(1)}\left(\alpha_{i}\right).
\end{aligned}
\end{equation}

Plugging (\ref{app_exp_mu_fisher_mse}) and (\ref{app_exp_mu_fisher_var}) into (\ref{app_exp_mu_fisher}), we have
\begin{equation}
\begin{aligned}
\label{app_exp_mu_fisher_final}
&\E \left[ (\vy - \vp)^T \mathcal{I}(\boldsymbol{\alpha})(\vy - \vp)\right] \\
= &\sum_{i=1}^{K} (y_i - \frac{\alpha_{i}}{\alpha_{0}} )^2 \psi^{(1)}\left(\alpha_{i}\right) + \frac{\alpha_{i}(\alpha_{0} - \alpha_{i})}{\alpha_{0}^2(\alpha_{0} + 1)} \psi^{(1)}\left(\alpha_{i}\right) 
= \sum_{i=1}^{K}  \left( (y_i - \frac{\alpha_{i}}{\alpha_{0}} )^2 + \frac{\alpha_{i}(\alpha_{0} - \alpha_{i})}{\alpha_{0}^2(\alpha_{0} + 1)} \right) \psi^{(1)}\left(\alpha_{i}\right)
\end{aligned}
\end{equation}
Furthermore, for the KL term, we have
\begin{equation}
\begin{aligned}
\label{app_exp_kl}
&\KL(Dir(\vp | \hat{\boldsymbol{\alpha}}) \Vert Dir(\vp|\boldsymbol{1}) ) \\
= &\log \Gamma(\sum_{k=1}^{K} \hat{\alpha}_{k}) - \log \Gamma(K) - \sum_{k=1}^{K} \log \Gamma(\hat{\alpha}_{k}) + \sum_{k=1}^{K}(\hat{\alpha}_{k} - 1) \left[ \psi(\hat{\alpha}_{k}) - \psi( \sum_{j=1}^{K} \hat{\alpha}_{j}) \right]
\end{aligned}
\end{equation}

Plugging (\ref{log_det_fisher}), (\ref{app_exp_mu_fisher_final}) and (\ref{app_exp_kl}) into (\ref{app_obj_func_exp}), we can obtain the final objective of Eq. \ref{obj_final}
$$
\begin{aligned}
&\E \left[- \log p(\vy|\vp, \boldsymbol{\alpha}, \sigma^2)\right] + \lambda \KL(Dir(\vp | \hat{\boldsymbol{\alpha}}) \Vert Dir(\vp|\boldsymbol{1}) ) \\
\propto & \underbrace{\sum_{i=1}^{K}  \left( (y_i - \frac{\alpha_{i}}{\alpha_{0}} )^2 + \frac{\alpha_{i}(\alpha_{0} - \alpha_{i})}{\alpha_{0}^2(\alpha_{0} + 1)} \right) \psi^{(1)}\left(\alpha_{i}\right)}_{\mathcal{L}_{i}^{\mathcal{I}\text{-MSE}}} -  \lambda_1 \underbrace{\left( \sum_{i=1}^{K} \log \psi^{(1)}(\alpha_{i}) + \log (1 - \sum_{i=1}^{K} \frac{\psi^{(1)}(\alpha_{0})}{\psi^{(1)}(\alpha_{i})}) \right)}_{\mathcal{L}_{i}^{|\mathcal{I}|}} \\
& + \lambda_2 \underbrace{\left(\log \Gamma(\sum_{k=1}^{K} \hat{\alpha}_{k}) - \log \Gamma(K) - \sum_{k=1}^{K} \log \Gamma(\hat{\alpha}_{k}) + \sum_{k=1}^{K}(\hat{\alpha}_{k} - 1) \left[ \psi(\hat{\alpha}_{k}) - \psi( \sum_{j=1}^{K} \hat{\alpha}_{j}) \right] \right)}_{\mathcal{L}_{i}^{\text{KL}}} 
.
\end{aligned}
$$

\subsection{Proof of Theorem \ref{pac}}
\label{app_proof_pac}

\begin{reptheorem}{pac}[\cite{germain2009pac, alquier2016properties, masegosa2020learning}]
Given a data distribution $\displaystyle P$ over $\mathcal{X} \times \mathcal{Y}$, a hypothesis set $\vtheta$, a prior distribution $\pi$ over $\vtheta$, for any $\delta \in (0,1]$, and $\lambda > 0$, with probability at least $1-\delta$ over samples $\displaystyle D \sim P^{n}$, we have for all posterior $\rho$,
$$
\displaystyle \E_{\rho(\vtheta)} [L(\vtheta)] \leq \E_{\rho(\vtheta)} [\hat{L}(\vtheta, D)] + \frac{1}{\lambda}\left[\KL(\rho \Vert \pi) + \log \frac{1}{\delta} + \Psi_{P, \pi}(\lambda, n) \right]
$$
where $\displaystyle \Psi_{P, \pi}(\lambda, n) = \log \E_{\pi(\vtheta)} \E_{D \sim P^{n}}\left[e^{\lambda(L(\vtheta)-\hat{L}(\vtheta, D))}\right] $.
\end{reptheorem}

\textit{Proof}. The Donsker-Varadhan's change of measure states that for any measurable function $\phi: \vtheta \rightarrow \mathbb{R}$, we have
$$
\displaystyle \E_{\rho(\vtheta)} [\phi(\vtheta)] \leq \KL(\rho \Vert \pi)+\log \E_{\pi(\vtheta)} [e^{\phi(\vtheta)}]
$$
Thus, with $\displaystyle \phi(\vtheta):=\lambda\left(L(\vtheta)-\hat{L}(\vtheta, D)\right)$, we obtain $\forall \rho$ on $\vtheta$:
$$
\begin{aligned}
\displaystyle \E_{\rho(\vtheta)} \left[\lambda\left(L(\vtheta)-\hat{L}(\vtheta, D)\right)\right] &= \lambda \left(\E_{\rho(\vtheta)} \left[L(\vtheta)\right] - \E_{\rho(\vtheta)} \left[\hat{L}(\vtheta, D)\right] \right) \\
&\leq \KL(\rho \Vert \pi)+\log \E_{\pi(\vtheta)} \left[e^{\lambda\left(L(\vtheta)-\hat{L}(\vtheta, D)\right)}\right]
\end{aligned}
$$
Next, we apply Markov's inequality on the random variable $\displaystyle  \zeta_{\pi}(D):=\E_{\pi(\vtheta)} \left[e^{\lambda\left(L(\vtheta)-\hat{L}(\vtheta, D)\right)}\right]$:
$$
\operatorname{Pr}\left(\zeta_{\pi}(D) \leq \frac{1}{\delta} \E_{D \sim P^{n}} \left[\zeta_{\pi}\left(D\right)\right]\right) \geq 1-\delta
$$
This implies that with probability at least $1-\delta$ over the choice of $\displaystyle D \sim P^{n}$, we have $\forall \rho$ on $\vtheta$:
$$
\displaystyle \operatorname{Pr}\left(\E_{\rho(\vtheta)} [L(\vtheta)] \leq \E_{\rho(\vtheta)} [\hat{L}(\vtheta, D)] + \frac{1}{\lambda}\left[\KL(\rho \Vert \pi) + \log \frac{1}{\delta} + \Psi_{P, \pi}(\lambda, n) \right]\right) \geq 1-\delta,
$$
where $\displaystyle \Psi_{P, \pi}(\lambda, n) = \log \E_{\pi(\vtheta)} \E_{D \sim P^{n}}\left[e^{\lambda(L(\vtheta)-\hat{L}(\vtheta, D))}\right] $.

\section{Derivations for Uncertainty Measures and Energy Distance}
\label{app_unc}

The Dirichlet distribution is parameterized by its concentration parameters $\boldsymbol{\alpha} = [\alpha_{1}, \cdots, \alpha_{K}]$, $\forall \alpha_{c}>0$, defined as:
$$
Dir(\vp|\boldsymbol{\alpha})=\frac{\Gamma\left(\alpha_{0}\right)}{\prod_{c=1}^{K} \Gamma\left(\alpha_{c}\right)} \prod_{c=1}^{K} p_{k}^{\alpha_{c}-1}, \alpha_{0}=\sum_{c=1}^{K} \alpha_{c}
$$
where $\vp \in \Delta^{K-1}$, and $\Gamma(\cdot)$ is the \textit{gamma} function. Note that the parameters $\boldsymbol{\alpha}$ is calculated by passing the input sample through the evidential network ($\boldsymbol{\alpha} = f_{\vtheta}(\vx) + 1 \in \mathbb{R}^{K}_{+}$). The following derivation is adapted from the Appendix of \cite{malinin2018predictive}.

\subsection{Expected Entropy of Dirichlet-based Uncertainty Models}
\label{app_unc_exp_ent}

The derivation of the expected entropy is as follows:
\begin{equation}
\begin{aligned}
\label{eq_exp_ent}
\E_{\vp \sim Dir(\boldsymbol{\alpha})} \left[\mathcal{H}[p(y \mid \vp)] \right] & = \int_{\mathcal{S}^{K-1}} Dir(\vp|\boldsymbol{\alpha}) \left( - \sum_{c=1}^{K} p_{c} \ln p_{c} \right) d \vp \\
& = - \sum_{c=1}^{K} \int_{\mathcal{S}^{K-1}} Dir(\vp|\boldsymbol{\alpha}) \left( p_{c} \ln p_{c} \right) d \vp \\
&= - \sum_{c=1}^{K} \int_{\mathcal{S}^{K-1}} \frac{\Gamma\left(\alpha_{0}\right)}{\prod_{k=1}^{K} \Gamma\left(\alpha_{k}\right)} \prod_{k=1}^{K} p_{k}^{\alpha_{k}-1} \left( p_{c} \ln p_{c} \right) d \vp \\
&= - \sum_{c=1}^{K} \int_{\mathcal{S}^{K-1}} \frac{\alpha_{c}}{\alpha_{0}}\frac{\Gamma\left(\alpha_{0} + 1 \right)}{\Gamma\left(\alpha_{c} + 1\right)\prod_{k=1, k \neq c}^{K} \Gamma\left(\alpha_{k}\right)} \prod_{k=1, k \neq c}^{K} p_{k}^{\alpha_{k}-1} p_{c}^{\alpha_{c}} \ln p_{c} d \vp \\
&= - \sum_{c=1}^{K} \frac{\alpha_{c}}{\alpha_{0}} \int_{\mathcal{S}^{K-1}} \E_{\vp \sim Dir([\alpha_{1}, \cdots, \alpha_{c-1}, \alpha_{c}+1, \alpha_{c+1}, \cdots, \alpha_{K}])}[\ln p_{c}]  d \vp \\
&= - \sum_{c=1}^{K} \frac{\alpha_{c}}{\alpha_{0}} \left( \psi(\alpha_c + 1)-\psi(\alpha_0 + 1) \right)
\end{aligned}
\end{equation}
The last third equation comes from the fact that $\Gamma(n)  = (n - 1)!$. Since the expected entropy captures the peaks of the output distribution $p(y \mid \vp)$, it is used to measure \textit{data uncertainty}. More specifically, lower entropy means that the model concentrates all probability mass on one class, while high entropy indicates that all $\vp$ generated by $Dir(\boldsymbol{\alpha})$ are more uniformly distributed. 

\subsection{Mutual Information of Dirichlet-based Uncertainty Models}
\label{app_unc_mi}

In the DBU models, the mutual information between the labels $y$ and the categorical $\vp$ can be deduced by computing the difference between the entropy of the expected distribution and the expected entropy of the distribution, which can be viewed as the difference between the total amount of uncertainty and the data uncertainty. 
\begin{equation}
\label{eq_mi}
\underbrace{I\left[y, \vp \mid \vx, \mathcal{D} \right]}_{\text{Distributional Uncertainty}} =\underbrace{\mathcal{H}\left[\E_{p(\vp \mid \vx, \mathcal{D})}[p(y \mid \vp)] \right]}_{\text {Total Uncertainty }}-\underbrace{\E_{p(\vp \mid \vx, \mathcal{D})}[\mathcal{H}[p(y \mid \vp)]]}_{\text{Expected Data Uncertainty}}. 
\end{equation}
Assuming that point estimate $p(\vtheta | \mathcal{D}) = \delta(\vtheta - \hat{\vtheta})  \Rightarrow p(\vp | \vx, \mathcal{D}) \approx p(\vp | \vx, \hat{\vtheta}) = Dir(\vp | \boldsymbol{\alpha})$ is sufficient given appropriate regularization and training data size, the mutual information can be simplified by
\begin{equation}
\begin{aligned}
\underbrace{I\left[y, \vp \mid \vx, \mathcal{D} \right]}_{\text{Distributional Uncertainty}} & \approx \underbrace{\mathcal{H}\left[\E_{\vp \sim Dir(\boldsymbol{\alpha})}[p(y \mid \vp)] \right]}_{\text {Total Uncertainty }}-\underbrace{\E_{\vp \sim Dir(\boldsymbol{\alpha})}[\mathcal{H}[p(y \mid \vp)]]}_{\text {Expected Data Uncertainty }} \\
&=-\sum_{c=1}^K \frac{\alpha_c}{\alpha_0} \ln \frac{\alpha_c}{\alpha_0} + \sum_{c=1}^K \frac{\alpha_c}{\alpha_0} \left( \psi\left(\alpha_c+1\right) - \psi\left(\alpha_0+1\right)\right) \\
&=-\sum_{c=1}^K \frac{\alpha_c}{\alpha_0}\left(\ln \frac{\alpha_c}{\alpha_0}-\psi\left(\alpha_c+1\right)+\psi\left(\alpha_0+1\right)\right)
\end{aligned}
\end{equation}
The second term in this derivation is from the results of \ref{app_unc_exp_ent}. The mutual information is often used to measure \textit{distributional uncertainty}, as it captures the uniform output distribution that excludes data uncertainty. High mutual information indicates a uniform distribution of expected categorical probability with low data uncertainty.

\subsection{Differential Entropy of Dirichlet-based Uncertainty Models}
\label{app_unc_diff_ent}

The derivation of the differential entropy is as follows:
\begin{equation}
\begin{aligned}
\label{eq_diff_ent}
\mathcal{H}\left[Dir(\vp \mid \boldsymbol{\alpha})\right] & = - \int_{\mathcal{S}^{K-1}} Dir(\vp|\boldsymbol{\alpha}) \ln Dir(\vp|\boldsymbol{\alpha}) d \vp \\
&= - \int_{\mathcal{S}^{K-1}} Dir(\vp|\boldsymbol{\alpha}) \left( \ln \Gamma\left(\alpha_{0}\right) - \sum_{c=1}^{K} \ln \Gamma\left(\alpha_{c}\right) + \sum_{c=1}^{K} (\alpha_{c}-1) \ln p_{k} \right) d \vp \\
&=\sum_{c=1}^K \ln \Gamma\left(\alpha_c\right)-\ln \Gamma\left(\alpha_0\right)-\sum_{c=1}^K\left(\alpha_c-1\right) \E_{\vp \sim Dir(\boldsymbol{\alpha})} [ \ln p_{c} ] \\
&=\sum_{c=1}^K \ln \Gamma\left(\alpha_c\right)-\ln \Gamma\left(\alpha_0\right)-\sum_{c=1}^K\left(\alpha_c-1\right) \left(\psi\left(\alpha_c\right)-\psi\left(\alpha_0\right)\right)
\end{aligned}
\end{equation}
The last equation comes from $\E_{\vp \sim Dir(\boldsymbol{\alpha})} [ \ln p_{c} ]  = \psi\left(\alpha_c\right)-\psi\left(\alpha_0\right)$. Lower entropy indicates the model yields a sharper distribution, while high entropy denotes a more uniform Dirichlet distribution. Thus, differential entropy is also a common measure of \textit{distributional uncertainty}.

\subsection{Energy Distance}
\label{app_energy}

Energy distance is a metric that measures the distance between the distributions of random vectors. Let $X$ and $Y$ be independent random vectors in $\mathbb{R}^{d}$, with cumulative distribution function (CDF) $F$ and $G$, respectively. The energy distance can be defined in terms of expected distances between the random vectors,
\begin{equation}
\label{eq_energy}
D(F, G) = \left( 2 \E \| X - Y \| - \E \| X - X^{\prime} \| - \E \| Y - Y^{\prime} \| \right)^{1/2},
\end{equation}
where $X$ and $X^{\prime}$ (resp. $Y$ and $Y^{\prime}$) are independent random variables whose probability distribution is $F$ (resp. $G$), $\| \cdot \|$ denotes the Euclidean norm. Energy distance is zero if and only if the distributions are identical.

\section{Experimental Details and Additional Results}
\label{app_exp}

\subsection{Datasets}
\label{app_exp_dataset}

\textbf{MNIST} \citep{lecun1998mnist} is a database of handwritten digits $0$ to $9$, consisting of a training set of $60,000$ examples and a test set of $10,000$ examples. Each input is composed of a $1 \times 28 \times 28$ tensor. We use $\textbf{(80\%, 20\%)}$ to split the training samples into training and validation sets. 
For OOD detection experiment, we use \textbf{KMNIST} \citep{clanuwat2018deep} and \textbf{FashionMNIST} \citep{xiao2017fashion} containing images of Japanese characters and images of clothes, respectively.

\textbf{CIFAR10} \citep{krizhevsky2009learning} consists of $60,000$ images in $10$ classes, including airplane, automobile, bird, cat, deer, dog, frog, horse, ship, truck. Among them, there are $50,000$ training images and $10,000$ test images. Each input is composed of a $3 \times 32 \times 32$ tensor. We use $\textbf{(95\%, 5\%)}$ to split the training samples into training and validation sets. 
Street View House Numbers (\textbf{SVHN}) dataset \citep{netzer2018street}, a dataset containing digital images, and \textbf{CIFAR100} are used for OOD detection.

\textbf{mini-ImageNet} \citep{vinyals2016matching} dataset was proposed for few-shot learning evaluation. It contains $100$ classes with $600$ samples of $84 \times 84$ color images per class. These $100$ classes are divided into $64$, $16$, and $20$ classes respectively for sampling tasks for meta-training, meta-validation, and meta-test. For the few-shot classification task, we evaluate $N$-way $K$-shot classification tasks for $N \in\{5, 10\}$ and $K \in \{1, 5, 20, 50\}$ and report the average accuracy($\%$,  top-1) and $95\%$ confidence interval over $10,000$ few-shot episodes on meta-test set. Each episode contains randomly sampled $N$ classes and $K$ samples per class for adaptation, and $\min(15, K)$ query samples per class for evaluation. For OOD detection experiments, we use Caltech-UCSD Birds (\textbf{CUB}) dataset \citep{wah2011caltech}, contains $11,788$ images of $200$ subcategories belonging to birds. Note that we randomly sample the same number of OOD examples as the query sample for OOD detection.

\textbf{tiered-ImageNet} \citep{ren2018meta} dataset is a larger subset of ILSVRC-12. Compared with the $100$ classes of mini-ImageNet, it contains $608$ classes (779,165 images), which are grouped into 34 higher-level nodes in the ImageNet human-curated hierarchy. The nodes are divided into $20, 6$, and $8$ disjoint sets of training, validation, and testing nodes, with the corresponding classes forming their respective meta-sets. We conduct the few-shot classification and OOD detection experiments for CUB on this dataset using the same settings as mini-ImageNet.

\subsection{Implementation details}
\label{app_exp_implem}

For the MNIST and CIFAR10 datasets, it is implemented by adapting the code provided by \cite{charpentier2020posterior}. Following \cite{charpentier2020posterior}, we use 3 convolutional layers with 3 dense layers and VGG16 \citep{simonyan2014very}, respectively. \textit{Softplus} is used in the last layer to get the non-negative output. We use a validation loss-based early termination strategy to train up to $200$ epochs with a batch size of $64$. The learning rate is set to $0.001$ for MNIST and FMNIST, $0.0005$ for CIFAR10. The coefficient $\lambda$ of -$|\mathcal{I}|$ is set by grid-search ($0.1, 0.05, 0.01, 0.005, 0.001$). The last chosen hyperparameter is $0.005$ for MNIST, $0.01$ for FMNIST and $0.05$ for CIFAR10.

For the mini-ImageNet and tiered-ImageNet few-shot classification experiments, it is implemented by adapting the
code provided by \citep{ghaffari2021importance}. More specifically, we use WideResNet-28-10 pre-trained backbone from \cite{yang2021free} as feature extractor to train the 1-layer classifier. \textit{Softplus} is used as the activation function to obtain the non-negative output. The  coefficient $\lambda$ is also set by grid-search on the meta-validation set. Table \ref{app_table_hyper} reports the last chosen hyperparameter for few-shot settings. Figure \ref{fig_fsl_mini_5w_fisher} shows the impact of $\lambda$ under $5$-way mini-ImageNet setting.

\begin{table}[h]
	\caption{List of hyperparameters for our approach.}
	\label{app_table_hyper}
	\begin{center}
	\resizebox{\textwidth}{!}{
		\begin{tabular}{lcccc|cccc}
			\toprule
			&\textbf{5way1shot} &\textbf{5way5shot}  &\textbf{5way20shot} &\textbf{5way50shot} &\textbf{10way1shot} &\textbf{10way5shot}  &\textbf{10way20shot} &\textbf{10way50shot} \\ 
                \midrule
                $\lambda$  & 0.01 & 0.05 & 0.005 & 0.01 & 0.1 & 0.01 & 0.01 & 0.01 \\
			\bottomrule
		\end{tabular}}
	\end{center}
 \vspace{-3mm}
\end{table}

\begin{table}[t]
	\caption{OOD detection results on (ID) CIFAR10 against (OOD) CIFAR100. Each experiment is run with 5 seeds. }
	\label{app_exp_cifar100_auroc}
	\begin{center}
	\resizebox{\textwidth}{!}{
		\begin{tabular}{ccccc|cccc}
		\toprule
            &\multicolumn{4}{c}{\textbf{AUPR}}  &\multicolumn{4}{c}{\textbf{AUROC}}  \\ 
			\cmidrule(lr){2-5}  \cmidrule(lr){6-9} 
			\textbf{Method}     &\textbf{Max. P}   &\textbf{$\alpha_0$}  &\textbf{D. Ent.}   &\textbf{M.I.} &\textbf{Max. P}   &\textbf{$\alpha_0$}  &\textbf{D. Ent.}   &\textbf{M.I.}\\ \midrule
                KL-PN & 61.41 $\pm$ 2.8 & 61.53 $\pm$ 3.4 & 60.21 $\pm$ 3.2 & 61.66 $\pm$ 3.4 & 57.89 $\pm$ 1.8 & 58.43 $\pm$ 2.6 & 55.94 $\pm$ 3.3 & 58.53 $\pm$ 2.6 \\
                RKL-PN & 55.42 $\pm$ 2.6 & 54.74 $\pm$ 2.8 & 55.40 $\pm$ 2.9 & 54.86 $\pm$ 2.9 & 54.24 $\pm$ 2.4 & 53.25 $\pm$ 2.9 & 54.15 $\pm$ 3.0 & 53.41 $\pm$ 3.0 \\
                PostN & 81.96 $\pm$ 0.8 & 82.06 $\pm$ 0.8 & 82.34 $\pm$ 0.8 & 78.64 $\pm$ 1.7 & 80.49 $\pm$ 0.9 & 81.17 $\pm$ 1.1 & 81.51 $\pm$ 1.0 & 80.22 $\pm$ 1.0 \\
                EDL & 84.30 $\pm$ 0.7 & 84.18 $\pm$ 0.7 & 84.32 $\pm$ 0.7 & 84.19 $\pm$ 0.7 & 80.96 $\pm$ 0.8 & 80.63 $\pm$ 1.0 & 80.99 $\pm$ 0.8 & 80.65 $\pm$ 1.0 \\
                \midrule
                \textbf{$\mathcal{I}$-EDL} & \textbf{85.35 $\pm$ 0.7} & \textbf{84.84 $\pm$ 0.6} & \textbf{85.40 $\pm$ 0.6} & \textbf{84.95 $\pm$ 0.7} & \textbf{83.55 $\pm$ 0.7} & \textbf{82.15 $\pm$ 0.5} & \textbf{83.69 $\pm$ 0.7} & \textbf{82.44 $\pm$ 0.5} \\
            \bottomrule
        \end{tabular}}
    \end{center}

	\caption{OOD detection results on (ID) MNIST against (OOD) KMNIST and FMNIST, (ID) CIFAR10 against (OOD) SVHN. Each experiment is run with 5 seeds. }
	\label{app_exp_auroc_other}
	\begin{center}
	\resizebox{\textwidth}{!}{
		\begin{tabular}{ccccc|cccc}
		\toprule
            &\multicolumn{4}{c}{\textbf{AUPR}}  &\multicolumn{4}{c}{\textbf{AUROC}}  \\ 
			\cmidrule(lr){2-5}  \cmidrule(lr){6-9} 
			\textbf{Method}     &\textbf{Max. P}   &\textbf{$\alpha_0$}  &\textbf{D. Ent.}   &\textbf{M.I.} &\textbf{Max. P}   &\textbf{$\alpha_0$}  &\textbf{D. Ent.}   &\textbf{M.I.}\\ \midrule
                &\multicolumn{8}{c}{\textbf{MNIST} $\rightarrow$ \textbf{KMNIST}} \\
                \cmidrule(lr){2-9}
                EDL & 97.02 $\pm$ 0.7 & 96.31 $\pm$ 0.2 & 96.92 $\pm$ 0.9 & 96.41 $\pm$ 1.8 & 96.59 $\pm$ 0.6 & 96.18 $\pm$ 1.3 & 96.49 $\pm$ 0.8 & 96.22 $\pm$ 1.3 \\
                \textbf{$\mathcal{I}$-EDL} & 98.34 $\pm$ 0.2 & 98.33 $\pm$ 0.2 & 98.34 $\pm$ 0.2 & 98.33 $\pm$ 0.2 & 98.00 $\pm$ 0.3 & 97.97 $\pm$ 0.3 & 97.99 $\pm$ 0.3 & 97.97 $\pm$ 0.3 \\
                \midrule
                &\multicolumn{8}{c}{\textbf{MNIST} $\rightarrow$ \textbf{FMNIST}} \\
                \cmidrule(lr){2-9}
                EDL & 98.10 $\pm$ 0.4 & 98.08 $\pm$ 0.4 & 98.10 $\pm$ 0.4 & 98.09 $\pm$ 0.4 & 97.39 $\pm$ 0.6 & 97.40 $\pm$ 0.5 & 97.48 $\pm$ 0.5 & 97.43 $\pm$ 0.5 \\
                \textbf{$\mathcal{I}$-EDL} & 98.89 $\pm$ 0.3 & 98.86 $\pm$ 0.3 & 98.89 $\pm$ 0.3 & 98.87 $\pm$ 0.2 & 98.49 $\pm$ 0.3 & 98.41 $\pm$ 0.4 & 98.48 $\pm$ 0.4 & 98.42 $\pm$ 0.4 \\
                \midrule
                &\multicolumn{8}{c}{\textbf{CIFAR10} $\rightarrow$ \textbf{SVHN}} \\
                \cmidrule(lr){2-9}
                EDL & 78.87 $\pm$ 3.5 & 79.12 $\pm$ 3.7 & 78.91 $\pm$ 3.5 & 79.11 $\pm$ 3.7 & 80.64 $\pm$ 4.2 & 81.06 $\pm$ 4.5 & 80.71 $\pm$ 4.3 & 81.05 $\pm$ 4.5 \\
                \textbf{$\mathcal{I}$-EDL} & 83.26 $\pm$ 2.4 & 82.96 $\pm$ 2.2 & 83.31 $\pm$ 2.5 & 83.06 $\pm$ 2.2 & 87.58 $\pm$ 2.0 & 86.79 $\pm$ 1.3 & 87.69 $\pm$ 2.1 & 87.01 $\pm$ 1.5 \\
            \bottomrule
        \end{tabular}}
    \end{center}

	\caption{Comparsion of label smoothing and $\mathcal{I}$-EDL on OOD detection AUPR of mini-ImageNet against CUB under $5$-way and $\{1, 5, 20, 50\}$-shot settings. Each experiment is run for over $10,000$ few-shot episodes.}
	\label{tab_label_smooth}
	\begin{center}
	\resizebox{\textwidth}{!}{
		\begin{tabular}{ccc|cc|cc|cc}
		\toprule
			&\multicolumn{2}{c}{\textbf{5-Way 1-shot}}  &\multicolumn{2}{c}{\textbf{5-Way 5-shot}} &\multicolumn{2}{c}{\textbf{5-way 20-shot}}  &\multicolumn{2}{c}{\textbf{5-Way 50-shot}} \\ 
             \cmidrule(lr){2-3}  \cmidrule(lr){4-5} \cmidrule(lr){6-7} \cmidrule(lr){8-9}
             \textbf{Method}     &\textbf{Max. P}   &\textbf{Ent.-$\alpha_0$}  &\textbf{Max. P}   &\textbf{Ent.-$\alpha_0$} &\textbf{Max. P}   &\textbf{Ent.-$\alpha_0$} &\textbf{Max. P}   &\textbf{Ent.-$\alpha_0$}\\ \midrule
			Label smoothing & 72.03 $\pm$ 0.2 & 73.00 $\pm$ 0.2 & 77.17 $\pm$ 0.2 & 77.11 $\pm$ 0.2 & 76.11 $\pm$ 0.2 & 75.35 $\pm$ 0.2 & 74.76 $\pm$ 0.2 & 73.86 $\pm$ 0.2 \\
            \textbf{$\mathcal{I}$-EDL} & 71.95 $\pm$ 0.2 & 74.79 $\pm$ 0.2 & 82.04 $\pm$ 0.2 & 82.49 $\pm$ 0.2 & 84.31 $\pm$ 0.2 & 85.42 $\pm$ 0.2 & 84.68 $\pm$ 0.2 & 84.91 $\pm$ 0.2 \\
            \midrule
            \textbf{$\Delta$} &\textbf{-0.08} & \textbf{1.79} & \textbf{4.87} & \textbf{5.38} & \textbf{8.20} & \textbf{10.07} & \textbf{9.92} & \textbf{11.05} \\
            \bottomrule
        \end{tabular}}
    \end{center}
\end{table}

\subsection{Additional Experimental Results on OOD detection}
\label{app_exp_results}

Table \ref{app_exp_cifar100_auroc} and \ref{app_exp_auroc_other} displays the AUPR and AUROC scores of OOD detection on CIFAR10 against CIFAR100 and SVHN, MNIST against KMNIST and FMNIST. Table \ref{tab_label_smooth} compares our method and label smoothing on the few-shot setting. We use maximum probability (Max.P) and entropy (Ent.) to measure uncertainty for label smoothing, and Max.P and $\alpha_0$ for $\mathcal{I}$-EDL. All of these results consistently demonstrate our proposed method's superior OOD detection performance.

\subsection{Additional Experimental Results on Few-shot Learning}
\label{app_exp_results_fsl}

\begin{table}[t]
	\caption{\textbf{AUROC} scores of OOD detection against CUB under $\{5, 10\}$-way and $\{1, 5, 20, 50\}$-shot settings. Each experiment is run for over $10,000$ few-shot episodes.}
	\label{app_tab_mini_all}
	\begin{center}
	\resizebox{\textwidth}{!}{
		\begin{tabular}{ccccc|cccc}
		\toprule
			&\multicolumn{4}{c}{\textbf{5-way 1-shot}}  &\multicolumn{4}{c}{\textbf{10-way 1-shot}} \\ 
             \cmidrule(lr){2-5} \cmidrule(lr){6-9}
             \textbf{Method}     &\textbf{Max.P}   &\textbf{$\alpha_0$} &\textbf{D. Ent.}   &\textbf{M.I.} &\textbf{Max.P}   &\textbf{$\alpha_0$} &\textbf{D. Ent.}   &\textbf{M.I.}\\ \midrule
			EDL & 61.88 $\pm$ 0.27 & 59.72 $\pm$ 0.31 & 63.60 $\pm$ 0.31 & 60.42 $\pm$ 0.30 & 55.83 $\pm$ 0.22 & 63.02 $\pm$ 0.29 & 63.06 $\pm$ 0.27 & 63.05 $\pm$ 0.29 \\
            \textbf{$\mathcal{I}$-EDL} & 67.46 $\pm$ 0.28 & 70.08 $\pm$ 0.30 & 69.51 $\pm$ 0.30 & 70.03 $\pm$ 0.30 & 68.81 $\pm$ 0.22 & 69.29 $\pm$ 0.22 & 68.80 $\pm$ 0.22 & 69.29 $\pm$ 0.22 \\
            \midrule
            \textbf{$\Delta$} & \textbf{5.58} & \textbf{10.36} & \textbf{5.91} & \textbf{9.61} & \textbf{12.98} & \textbf{6.27} & \textbf{5.74} & \textbf{6.24} \\
                \midrule
                \midrule
                &\multicolumn{4}{c}{\textbf{5-way 5-shot}}  &\multicolumn{4}{c}{\textbf{10-way 5-shot}} \\ 
             \cmidrule(lr){2-5} \cmidrule(lr){6-9}
             \textbf{Method}     &\textbf{Max.P}   &\textbf{$\alpha_0$} &\textbf{D. Ent.}   &\textbf{M.I.} &\textbf{Max.P}   &\textbf{$\alpha_0$} &\textbf{D. Ent.}   &\textbf{M.I.}\\ \midrule
			EDL & 69.71 $\pm$ 0.25 & 72.01 $\pm$ 0.33 & 72.97 $\pm$ 0.29 & 72.14 $\pm$ 0.32 & 68.59 $\pm$ 0.23 & 73.23 $\pm$ 0.23 & 72.70 $\pm$ 0.23 & 73.16 $\pm$ 0.23 \\
            \textbf{$\mathcal{I}$-EDL} & 79.33 $\pm$ 0.22 & 79.81 $\pm$ 0.22 & 79.60 $\pm$ 0.22 & 79.79 $\pm$ 0.22 & 79.34 $\pm$ 0.23 & 80.29 $\pm$ 0.22 & 78.36 $\pm$ 0.20 & 79.91 $\pm$ 0.21  \\
            \midrule
            \textbf{$\Delta$} & \textbf{9.62} & \textbf{7.80} & \textbf{6.63} & \textbf{7.65} & \textbf{10.75} & \textbf{7.06} & \textbf{5.66} & \textbf{6.75} \\
                \midrule
            \midrule
            &\multicolumn{4}{c}{\textbf{5-way 20-shot}}  &\multicolumn{4}{c}{\textbf{10-way 20-shot}} \\ 
             \cmidrule(lr){2-5} \cmidrule(lr){6-9}
             \textbf{Method}     &\textbf{Max.P}   &\textbf{$\alpha_0$} &\textbf{D. Ent.}   &\textbf{M.I.} &\textbf{Max.P}   &\textbf{$\alpha_0$} &\textbf{D. Ent.}   &\textbf{M.I.}\\ \midrule
			EDL & 76.59 $\pm$ 0.24 & 76.16 $\pm$ 0.29 & 76.75 $\pm$ 0.27 & 76.18 $\pm$ 0.29 & 71.69 $\pm$ 0.19 & 74.08 $\pm$ 0.20 & 73.88 $\pm$ 0.19 & 74.05 $\pm$ 0.20 \\
            \textbf{$\mathcal{I}$-EDL} & 82.04 $\pm$ 0.21 & 83.38 $\pm$ 0.22 & 83.00 $\pm$ 0.21 & 83.29 $\pm$ 0.21 & 79.66 $\pm$ 0.16 & 80.74 $\pm$ 0.16 & 80.07 $\pm$ 0.15 & 80.61 $\pm$ 0.16 \\
            \midrule
            \textbf{$\Delta$} & \textbf{5.45} & \textbf{7.22} & \textbf{6.25} & \textbf{7.11} & \textbf{7.96} & \textbf{6.66} & \textbf{6.19} & \textbf{6.56}  \\
                \midrule
            \midrule
            &\multicolumn{4}{c}{\textbf{5-way 50-shot}}  &\multicolumn{4}{c}{\textbf{10-way 50-shot}} \\ 
             \cmidrule(lr){2-5} \cmidrule(lr){6-9}
             \textbf{Method}     &\textbf{Max.P}   &\textbf{$\alpha_0$} &\textbf{D. Ent.}   &\textbf{M.I.} &\textbf{Max.P}   &\textbf{$\alpha_0$} &\textbf{D. Ent.}   &\textbf{M.I.}\\ \midrule
			EDL & 79.30 $\pm$ 0.20 & 78.74 $\pm$ 0.24 & 79.21 $\pm$ 0.22 & 78.76 $\pm$ 0.24 & 73.67 $\pm$ 0.17 & 74.43 $\pm$ 0.18 & 74.32 $\pm$ 0.17 & 74.40 $\pm$ 0.18 \\
            \textbf{$\mathcal{I}$-EDL} & 82.55 $\pm$ 0.17 & 83.17 $\pm$ 0.19 & 83.29 $\pm$ 0.18 & 83.20 $\pm$ 0.19 & 77.39 $\pm$ 0.15 & 77.80 $\pm$ 0.16 & 77.72 $\pm$ 0.15 & 77.78 $\pm$ 0.16 \\
            \midrule
            \textbf{$\Delta$} & \textbf{3.25} & \textbf{4.43} & \textbf{4.08} & \textbf{4.44} & \textbf{3.72} & \textbf{3.37} & \textbf{3.40} & \textbf{3.38} \\
            \bottomrule
        \end{tabular}}
    \end{center}
\end{table}

\begin{figure}[t]
    \centering
    \includegraphics[width=1.0\textwidth]{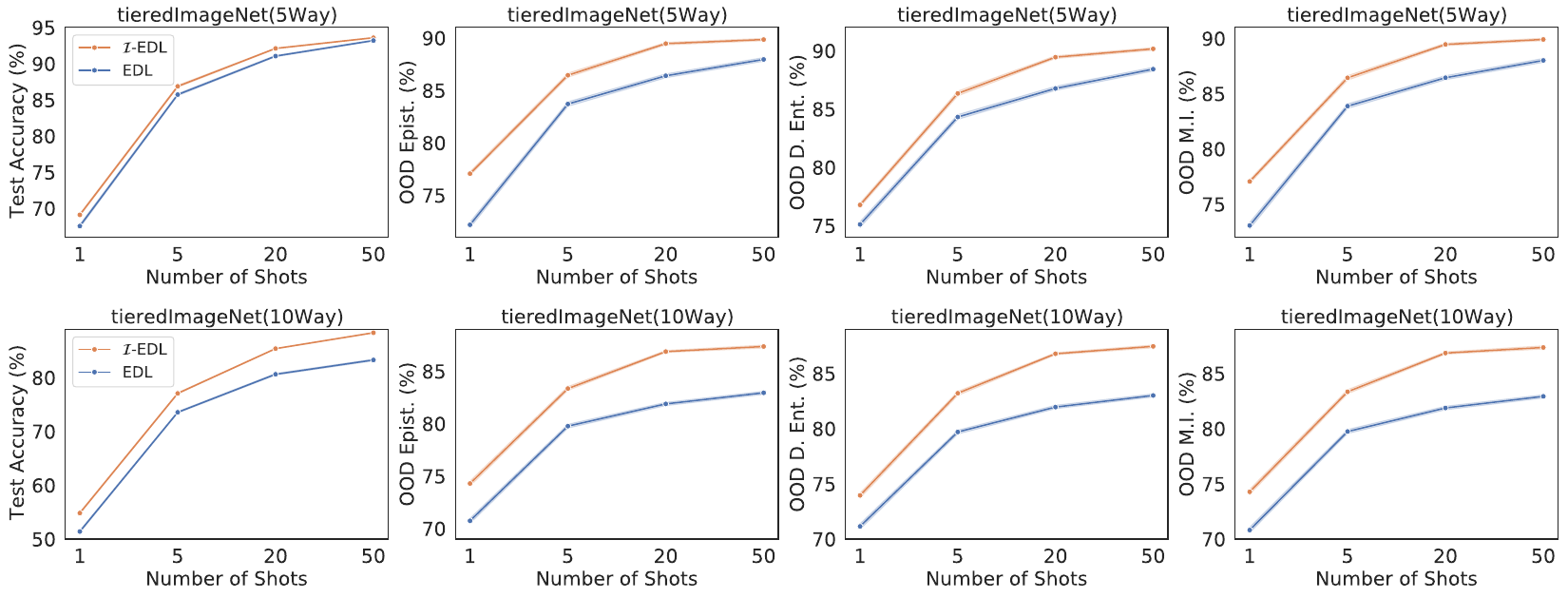}
    \caption{Results on tiered-ImageNet $\{5, 10\}$-way $\{1, 5, 20, 50\}$-shot classification (\textbf{Accuracy}) and OOD detection against CUB (\textbf{AUPR}). The error bars are almost non-existent (i.e., less than $0.03\%$), since over $10,000$ trials were performed for each point. $\mathcal{I}$-EDL produces statistically significant improvements over a wide range of the number of shots. }
    \label{fig_fsl_tiered}
\end{figure}

Table \ref{app_tab_mini_all} shows the AUROC scores of OOD detection under $\{5, 10\}$-way $\{1, 5, 20, 50\}$-shot of mini-ImageNet. For all the $N$-way $K$-shot tasks, AUROC with Max.P, $\alpha_0$, D.Ent, and M.I. show impressive improvements on $\mathcal{I}$-EDL over EDL. For example, the improvements of OOD detection after using $\mathcal{I}$-EDL are all above $\textbf{3.37\%}$, especially the improvement of Max.P-based OOD detection is up to $\textbf{12.98\%}$ under $10$-way $1$-shot. The experimental results of tiered-ImageNet are shown in Figure \ref{fig_fsl_tiered}. We can observe the same experimental results as mini-ImageNet, which indicates that our method improves the performance of OOD detection, especially in the more challenging few-shot setting.

\subsection{Additional Ablation Study}
\label{app_exp_hyper}

\begin{figure}[t]
    \centering
    \includegraphics[width=1.0\textwidth]{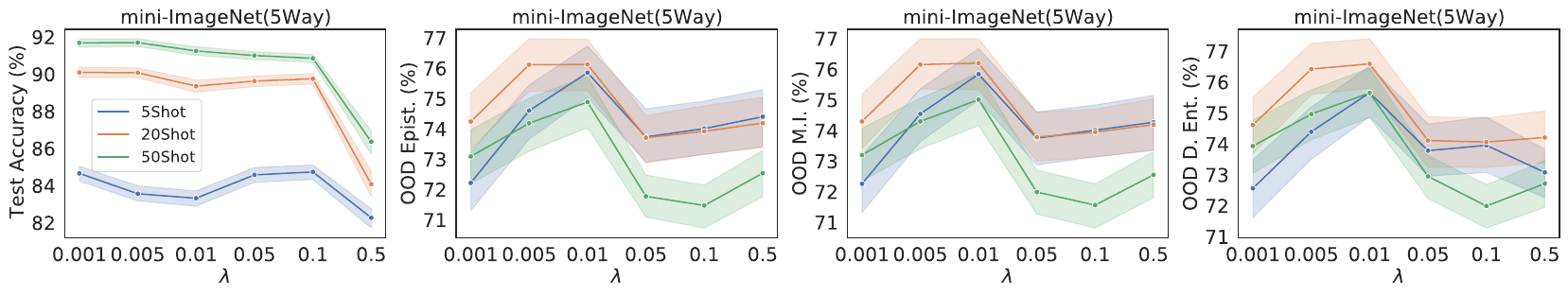}
    \vspace{-3mm}
    \caption{The impact of coefficient $\lambda$ of the negative log determinant of the FIM (-$|\mathcal{I}|$) under $5$-way mini-ImageNet classification and OOD detection experiments.}
    \label{fig_fsl_mini_5w_fisher}
    \vspace{-3mm}
\end{figure}

Figure \ref{fig_fsl_mini_5w_fisher} shows the effect of coefficient $\lambda$ of the negative log determinant of FIM (-$|\mathcal{I}|$) under mini-ImageNet classification and OOD detection experiments. We plot the results under $5$-way $\{5, 20, 50\}$-shot. It can be observed that the best coefficients for OOD detection based on different uncertainty measures show consistency, but not with the best coefficients for accuracy. More specifically, the best coefficients for OOD detection are all around $0.01$, but accuracy prefers $0.001$. This is a non-trivial problem because it involves a multi-objective optimization problem. In this work, the coefficient is ultimately a compromise choice that combines the performance of image classification and OOD detection. Whether there is a better way to optimize multiple objectives at the same time remains to be explored in the future.

\subsection{Additional Analysis of Uncertainty estimation}
\label{app_exp_unc}

\begin{figure}[t]
    \centering
    \includegraphics[width=0.98\textwidth]{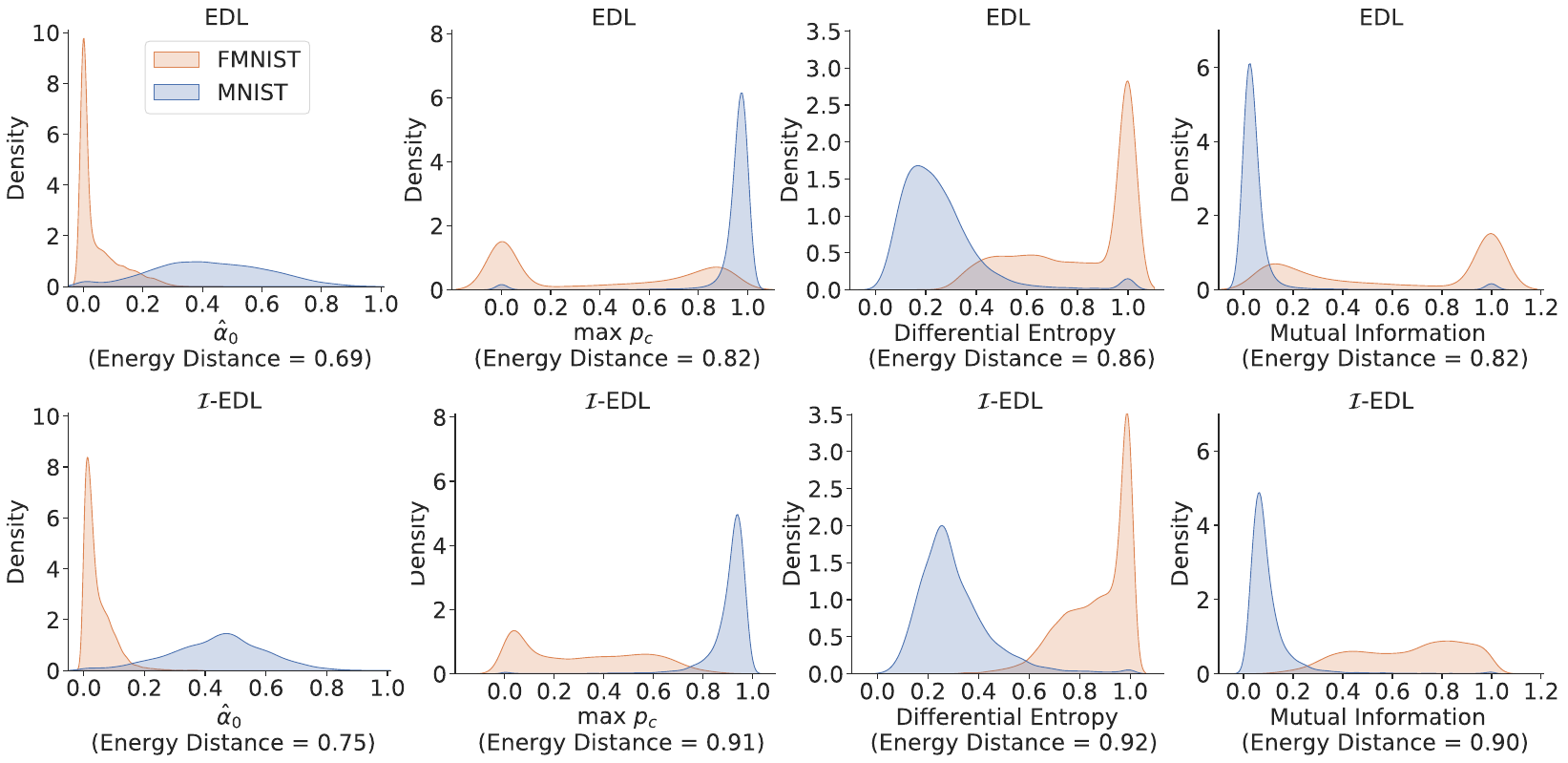}
    \vspace{-3mm}
    \caption{Uncertainty representation for ID (MNIST) and OOD (FMNIST). }
    \label{fig_density_fmnist}
    \vspace{-3mm}
\end{figure}

\begin{figure}[t]
    \centering
    \includegraphics[width=0.98\textwidth]{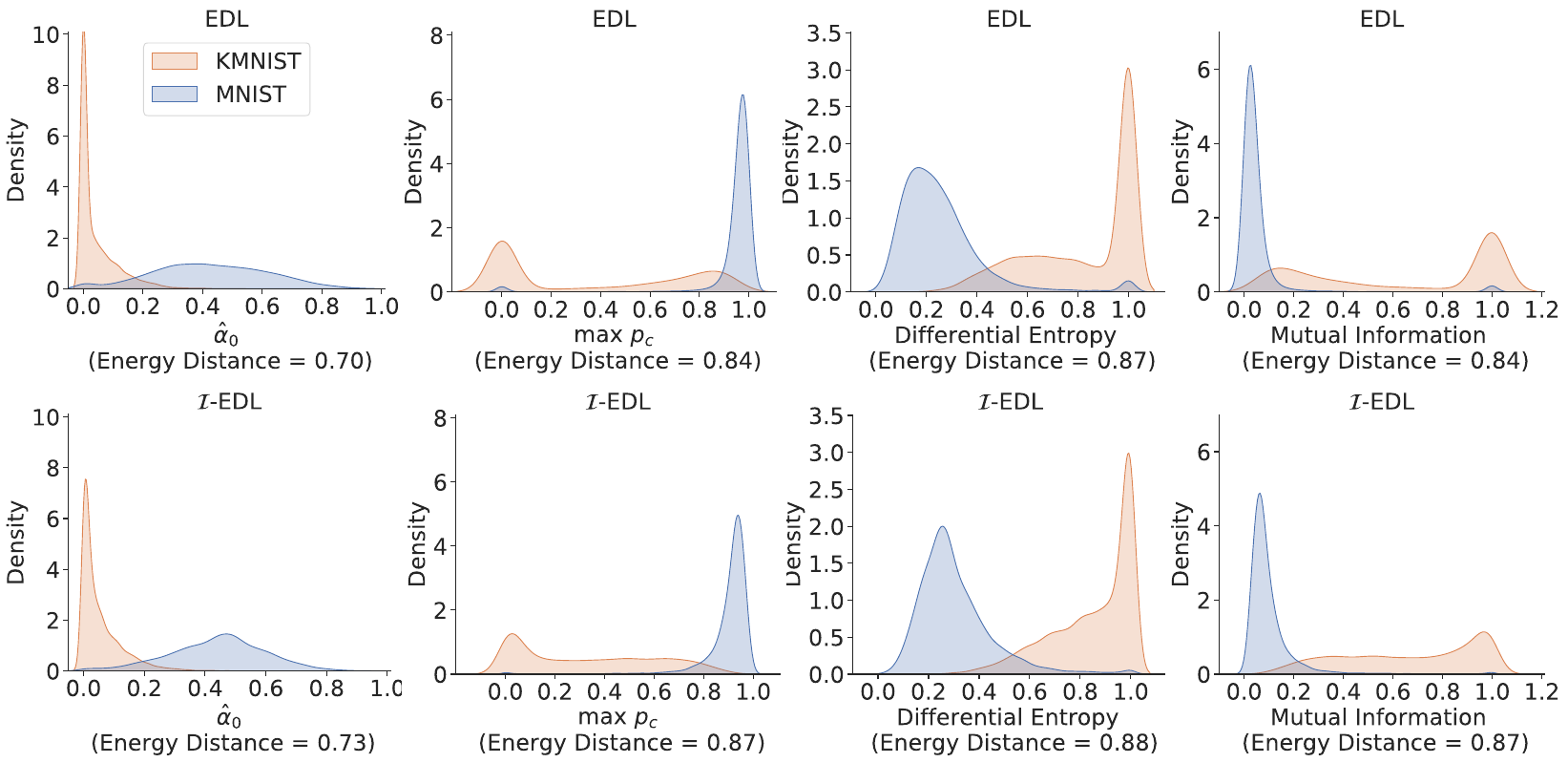}
    \vspace{-3mm}
    \caption{Uncertainty representation for ID (MNIST) and OOD (KMNIST). }
    \label{fig_density_kmnist}
    \vspace{-3mm}
\end{figure}

Figure \ref{fig_density_fmnist} and \ref{fig_density_kmnist} shows density plots of the normalized uncertainty measures for MNIST vs FMNIST, and MNIST vs KMNIST, respectively. The uncertainty measures include precision $\alpha_0$, $\max_{c} p_c$, differential entropy and mutual information. We normalize each uncertainty value $u_c$ by $u_c = (u_c - \min_{i} u_i) / (\max_{i} u_i - \min_{i} u_i)$. We also report the energy distance of two distributions, with higher values indicating more separability. It can be observed that $\mathcal{I}$-EDL produces sharper prediction peaks than EDL in the in-distribution (MNIST) region. Although not using OOD data, our method also makes the uncertainty of OOD data more aggregated.

\end{document}